\documentclass[10pt,twocolumn,letterpaper]{article}

\usepackage{cvpr}
\usepackage{times}
\usepackage{epsfig}
\usepackage{graphicx}
\usepackage{amsmath}
\usepackage{amssymb}
\usepackage{makecell}

\usepackage{hyperref}
\hypersetup{pagebackref=true,breaklinks=true,letterpaper=true,colorlinks,bookmarks=false}

\cvprfinalcopy % *** Uncomment this line for the final submission

\def\cvprPaperID{1476} % *** Enter the CVPR Paper ID here

\ifcvprfinal\fi
\begin{document}

%%%%%%%%% TITLE
\title{Polarized Reflection Removal with Perfect Alignment in the Wild}

% \author{Chenyang Lei\\
% HKUST\\
% Institution1 address\\
% {\tt\small firstauthor@i1.org}
% \and
% Xuhua Huang\\
% HKUST\\
% Institution2 address\\
% {\tt\small secondauthor@i2.org}
% \and
% Mengdi Zhang\\
% HKUST\\
% \and
% Qiong Yan\\
% Sensetime\\
% \and
% Wenxiu Sun\\
% Sensetime\\
% \and
% Qifeng Chen\\
% HKUST\\
% }
\author{Chenyang Lei$^{1}$ \quad Xuhua Huang$^1$ \quad  Mengdi Zhang$^1$ \quad  Qiong Yan$^2$ \quad  Wenxiu Sun$^2$ \quad  Qifeng Chen$^1$\\
$^1$HKUST  \qquad $^2$Sensetime\\
}

% For a paper whose authors are all at the same institution,
% omit the following lines up until the closing ``}''.
% Additional authors and addresses can be added with ``\and'',
% just like the second author.
% To save space, use either the email address or home page, not both

\maketitle
% \thispagestyle{empty}

%%%%%%%%% ABSTRACT
\begin{abstract}
We present a novel formulation to removing reflection from polarized images in the wild. We first identify the misalignment issues of existing reflection removal datasets where the collected reflection-free images are not perfectly aligned with input mixed images due to glass refraction. Then we build a new dataset with more than 100 types of glass in which obtained transmission images are perfectly aligned with input mixed images. Second, capitalizing on the special relationship between reflection and polarized light, we propose a polarized reflection removal model with a two-stage architecture. In addition, we design a novel perceptual NCC loss that can improve the performance of reflection removal and general image decomposition tasks. We conduct extensive experiments, and results suggest that our model outperforms state-of-the-art methods on reflection removal.

\end{abstract}

%%%%%%%%% BODY TEXT
\section{Introduction}
%What is the problem? Why is it important?
% Reflection on the glass is usually undesirable in photography or computer vision tasks. In most cases, we want to have only the light behind the glass passed through,
% % the clean image pass through the glass, 
% which is transmission. Reflection, light reflected from the front, will have noticeable bad influence on the quality of photos or the performance of computer vision tasks, such as object detection, image classification and image matching. Reflection removal is therefore beneficial in these scenarios.
% % Removing reflection can not only improve the photo quality but also enhance the performance of these algorithms.

It is often desirable to remove glass reflection as it may contaminate the visual quality of a photograph. Reflection separation is also arguably important for robots to work robustly in the real world as the content in reflection usually does not exist in the viewing frustum of a camera. One intriguing property of reflection is that reflected light is often polarized, which may facilitate reflection removal. In this paper, we study reflection removal with polarized sensors by designing a customized deep learning framework.

\begin{figure}
\centering
\includegraphics[width=0.9\linewidth]{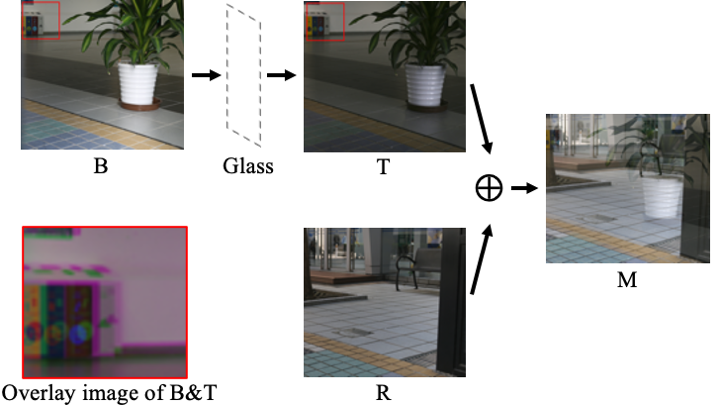}
\caption{The image formation process of the mixed image $M$. Due to refraction, background $B$ dims and shifts after passing the glass and forms transmission $T$, resulting in intensity discrepancy and spatial misalignment between $B$ and $T$. Reflection $R$ from the glass surface is linearly added to $T$ to form $M$ in raw data space. Therefore, we can obtain $T$ by computing $T=M-R$. 
%Background and transmission are different in real world. $T$ is always darker than $B$ as some light is reflected or absorbed by glass, and there is a spatial shift between $T$ and $B$ due to refraction.  Previous methods ignore the difference between $B$ and $T$, which introduces the problems of intensity decay, color distortion, and \textbf{misalignment}. To solve these problems, we propose the $T=M-R$ method to get the transmission.
}
\label{fig:M-R}
\end{figure}

%Why is the problem hard? What makes it challenging?
%这里要把我们的formulation稍微简单说一下，什么是background，什么是transmission,再过几段详细说明他们的问题；
An image with reflection is a mixture of reflection and transmission, as shown in Fig. \ref{fig:M-R}. In raw data space, the mixed image $M$ can be formulated as
\begin{align}
\label{eq:MTR}
    M = T + R,
\end{align}
where $T$ and $R$ are transmission and reflection, respectively. We name the light behind glass as background $B$ and the light that passes through glass as transmission $T$. Although most prior work treats $B$ as the same as $T$~\cite{zhang2018single,wei2019single_ERR}, we argue that $T$ and $B$ are different. $T$ is darker than $B$ as some light is reflected or absorbed by glass, and there is a spatial shift between $T$ and $B$ due to refraction. 

A common issue of many existing reflection removal methods~\cite{zhang2018single,Yang_2019_CVPR,fan2017generic,eccv18refrmv_BDN,shih2015reflection} is that strict assumptions are imposed on reflection. These assumptions make previous methods work well in special cases but fail in many others. For example, many works assume reflection images are out of focus \cite{fan2017generic,zhang2018single}. As a result, these approaches may not remove reflection properly when the reflection is sharp and strong. Another prior assumption is on ghost cues~\cite{shih2015reflection} that result from multiple reflections inside a thick glass. However, ghost cues do not exist in thin glass.

The lack of diverse and high quality real-world data is another challenging issue. Zhang et al.~\cite{zhang2018single} and Wei et al.~\cite{wei2019single_ERR} have collected a small set of real-world data where only background images (in Fig. \ref{fig:M-R} and Fig. \ref{fig:PriorData}) are captured as the ground-truth transmission images. However, background images are not perfectly aligned with the mixed images $M$ due to refraction and also have the problem of intensity decay ($T$ appears darker than $B$) and color distortion (colored glass). Misalignment introduces great challenges in training a machine learning model~\cite{wei2019single_ERR} and the intensity difference makes it even more difficult. Moreover, since the type of reflection depends on the glass type and only one type of glass is used to collect data, the models trained on these data cannot generalize well to other types of glass.

\begin{figure}[t]
\centering
\begin{tabular}{@{}c@{\hspace{1mm}}c@{\hspace{1mm}}c@{}}
\includegraphics[width=0.32\linewidth]{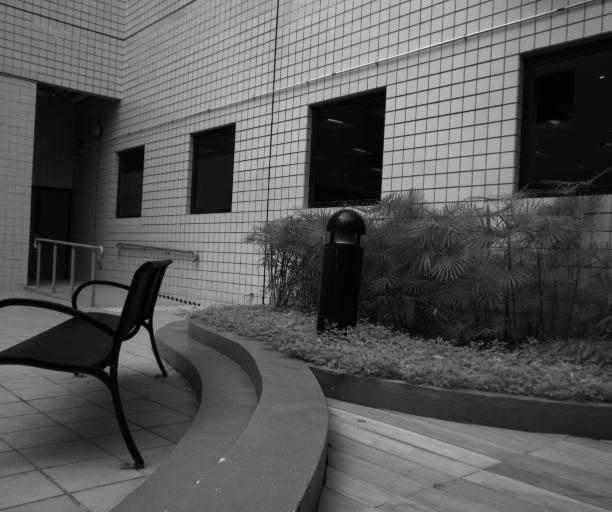}&
\includegraphics[width=0.32\linewidth]{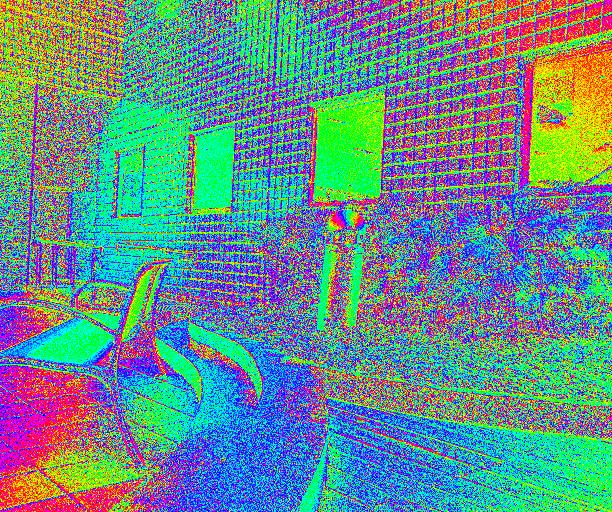}&
\includegraphics[width=0.32\linewidth]{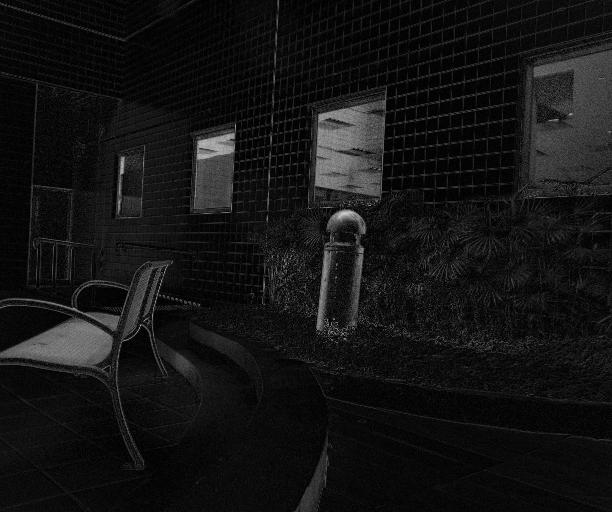}\\
Raw & $\phi$ & $\rho$\\
\end{tabular}
%\vspace{1mm}
\caption{The visualization of polarization information. Polarization exists inside and outside the glass area. $\phi$ is the angle of polarization and visualized by hue in HSV space. $\rho$ is the degree of polarization and visualized by intensity.
%If the pixel is completely black, then the $\rho$ equals to 0. in The raw image is taken by a novel polarization sensor. We visualize the angle using HSV space where $\phi$ decides the hue.
}
\label{fig:Polarization information}
\end{figure}

To be able to relax the assumptions about the appearance of reflection, we leverage polarization that inherently exists in almost all reflected light. Fig. \ref{fig:Polarization information} shows an example polarized image. Existing works based on polarization often impose strict assumptions. A common one is that all light sources are unpolarized~\cite{atkinson2006recovery}, which is easily violated in the real world because reflection happens in different types of surfaces in addition to glass and polarized or partially polarized light source exists commonly, such as the LED light. As can be seen in Fig. \ref{fig:Polarization information}, polarization exists both inside and outside the glass. We cannot solely rely on this information. To rule out the case that polarization also happens in transmission image, our work in this paper removes this assumption. Therefore, our method is more general and applicable to more scenarios.

 %Problem2: Real world data, Our method: how to solve this problem?

To ensure the diversity and quality of real-world data, we propose a new data collection pipeline called M-R based on the principle that raw image space is linear. We capture $M$ and $R$ only and obtain the transmission through $T=M-R$. Note that we capture the raw sensor data so that Eq.~\ref{eq:MTR} holds. Our formulation is physically faithful to image formation and eases the process of data collection. We show that with our novel M-R pipeline, it is easy to capture reflection caused by a diverse set of glass. We use the M-R pipeline to build a real-world polarization dataset collected by a novel polarization sensor for reflection removal.

With the collected dataset, we propose a two-stage framework for reflection removal from polarized images. Our approach firstly estimates reflection, with which it infers the transmission image secondly. Our PNCC (perceptual NCC) loss is used to minimize the similarity between the output reflection and transmission. Experiments demonstrate that our method achieves state-of-the-art performances on various metrics. The ablation study shows that our approach benefits from polarized data, PNCC, and the two-stage framework design. Our contributions are summarized as follows:
\begin{itemize}

    \item We observe two important factors for the task of reflection removal: 1) the difference between transmission $T$ and background $B$ is noticeable. 2) the linearity from reflection to mixed image holds perfectly on raw data. 
    
    \item We design a new data collection pipeline called M-R, which helps us collect diverse real-world data with perfect alignment by utilizing glass in the real world.
    
    \item We propose a deep learning method for reflection removal based on polarization data. Our method does not impose any assumption on the appearance of reflection. A two-stage framework is adopted to get better performance. We design a PNCC loss, which can be applied to many image decomposition tasks. Experiments show that our method outperforms all state-of-the-art methods and has better generalization.
    
\end{itemize}

 %Problem2: Real world data. %Problem2: Real world data, why is the problem important?
 %(1) Without real world data, they will fail in real world cases
 %(2) But their collection method has problem

%What does our paper contribute? What is the key idea? What is the magic trick? What is the new insight or technique that enables us to advance the frontier?
%(1) Less Assumption-> polarization information
%(2) New formulation for the glass model
%(3) PNCC Loss

\begin{figure*}[h]
\centering
\begin{tabular}{@{}c@{\hspace{1mm}}c@{\hspace{1mm}}c@{\hspace{1mm}}c@{\hspace{1mm}}c@{\hspace{5mm}}c@{\hspace{1mm}}c@{\hspace{1mm}}c@{}}

% \rotatebox{90}{\small \hspace{4mm}After ISP}
&\includegraphics[width=0.13\linewidth]{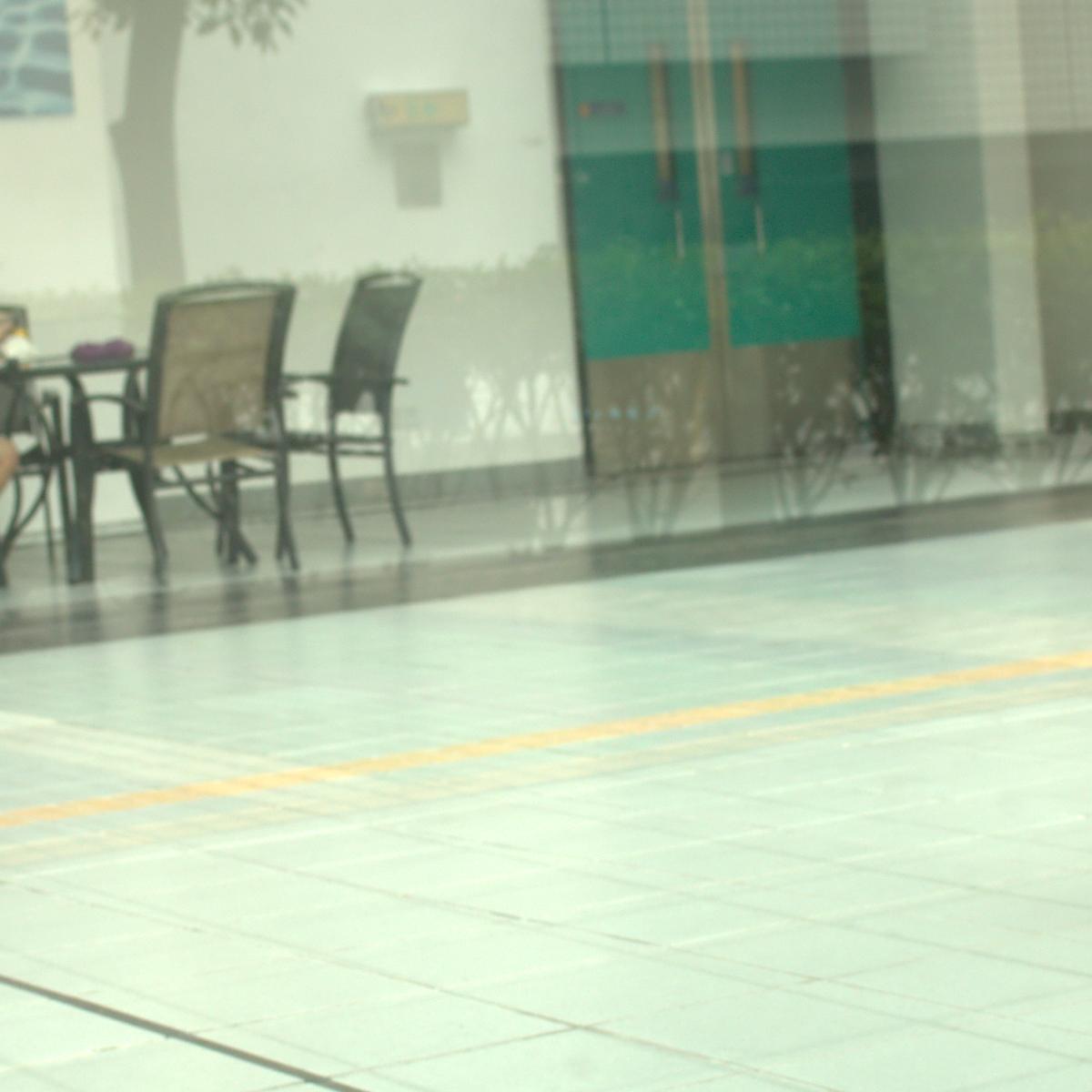}&
\includegraphics[width=0.13\linewidth]{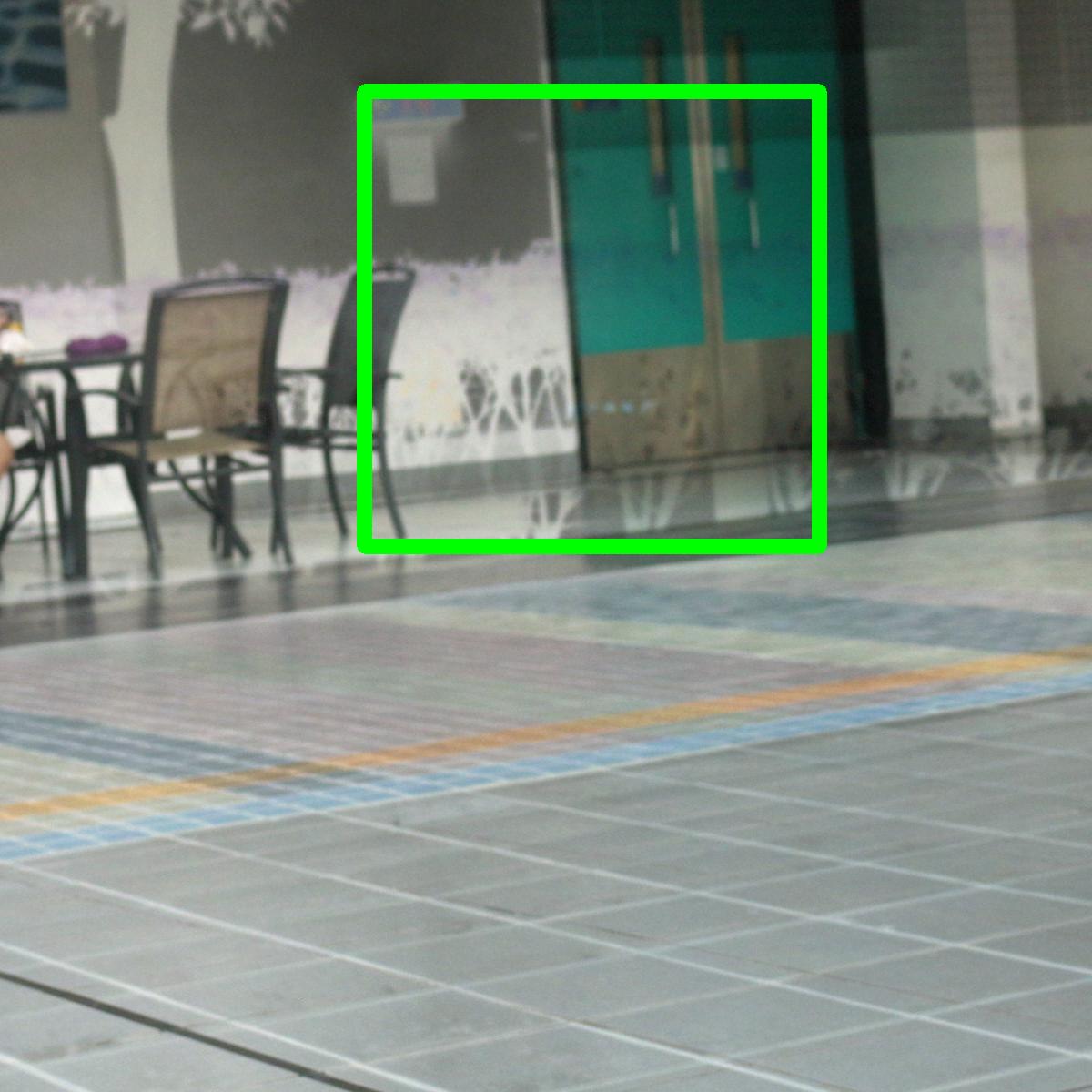}&
\includegraphics[width=0.13\linewidth]{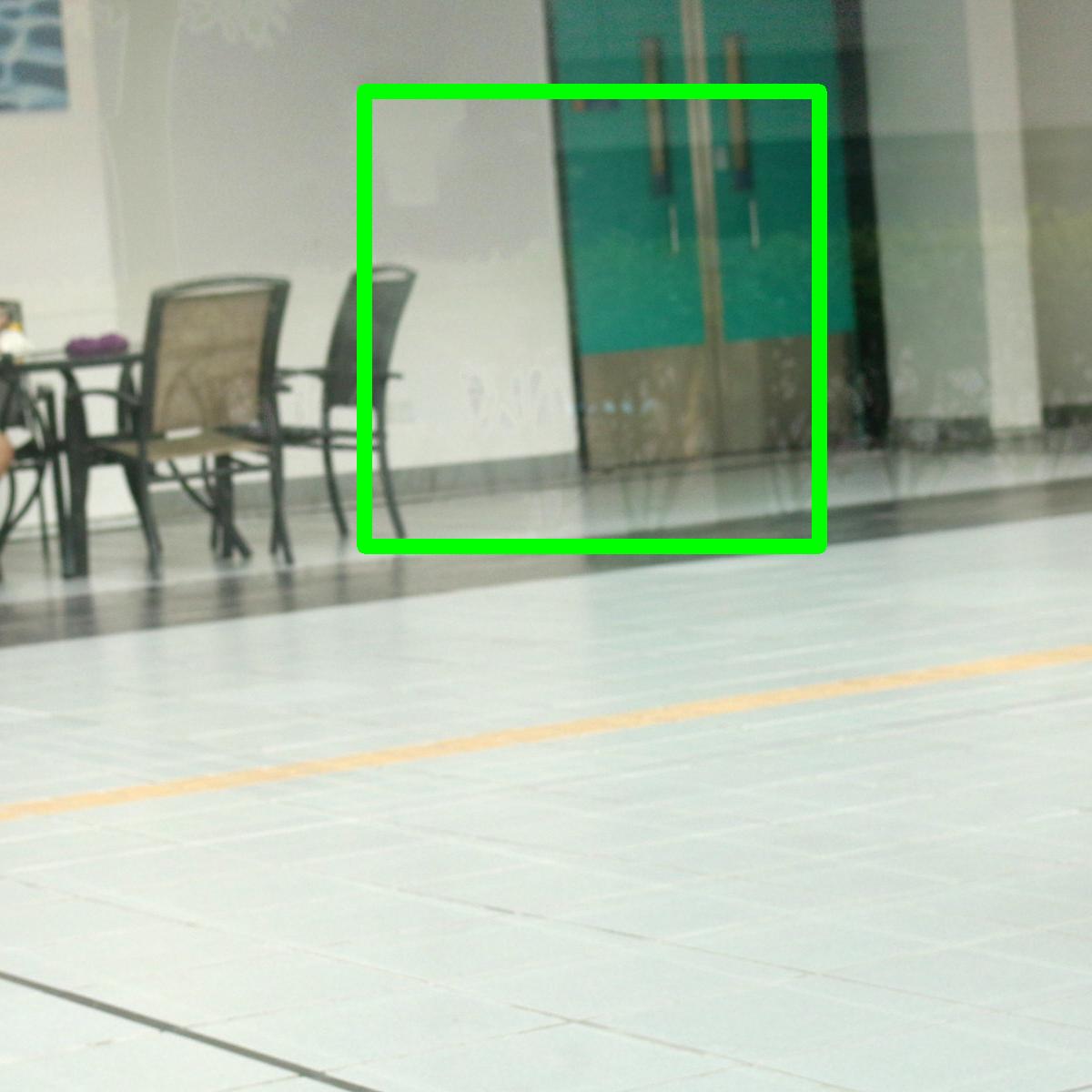}&
\includegraphics[width=0.13\linewidth]{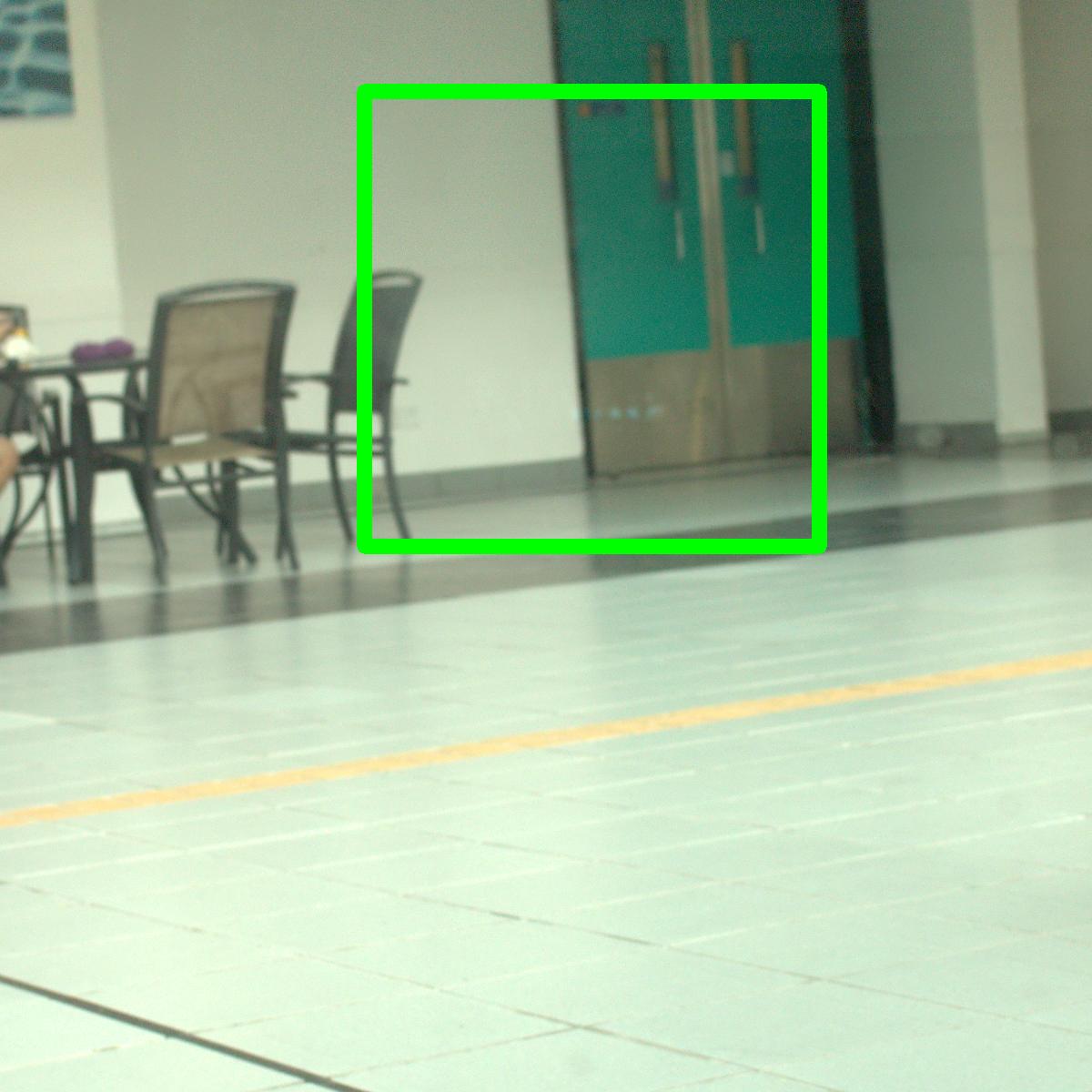}&
\includegraphics[width=0.13\linewidth]{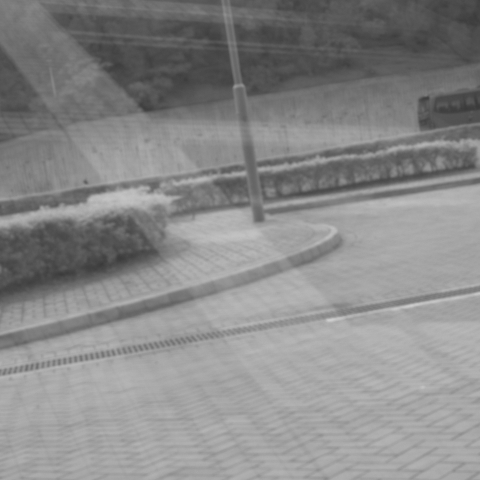}&
\includegraphics[width=0.13\linewidth]{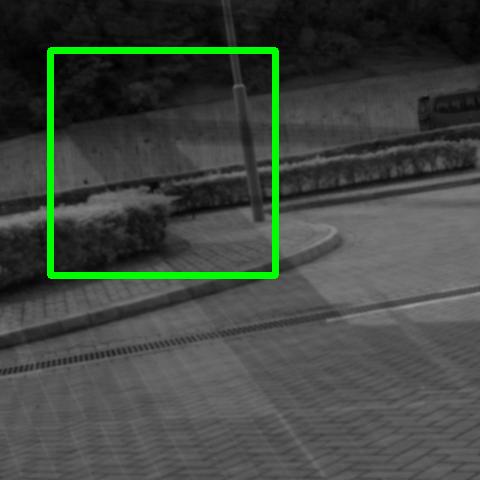}&
\includegraphics[width=0.13\linewidth]{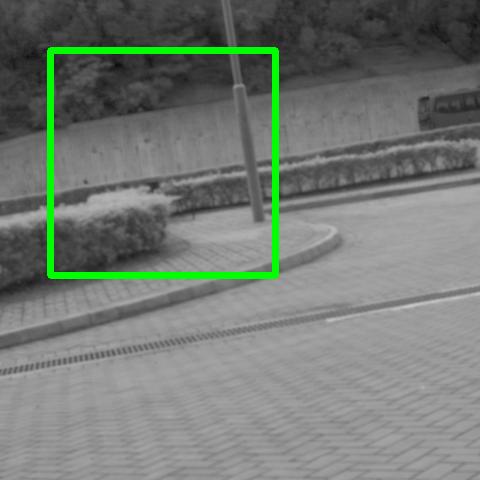}\\

&RGB M & ISP M-R & Gamma M-R &Raw M-R& Pol M& ISP M-R& Raw M-R\\

% \rotatebox{90}{\small \hspace{6mm}Raw}
&\includegraphics[width=0.13\linewidth]{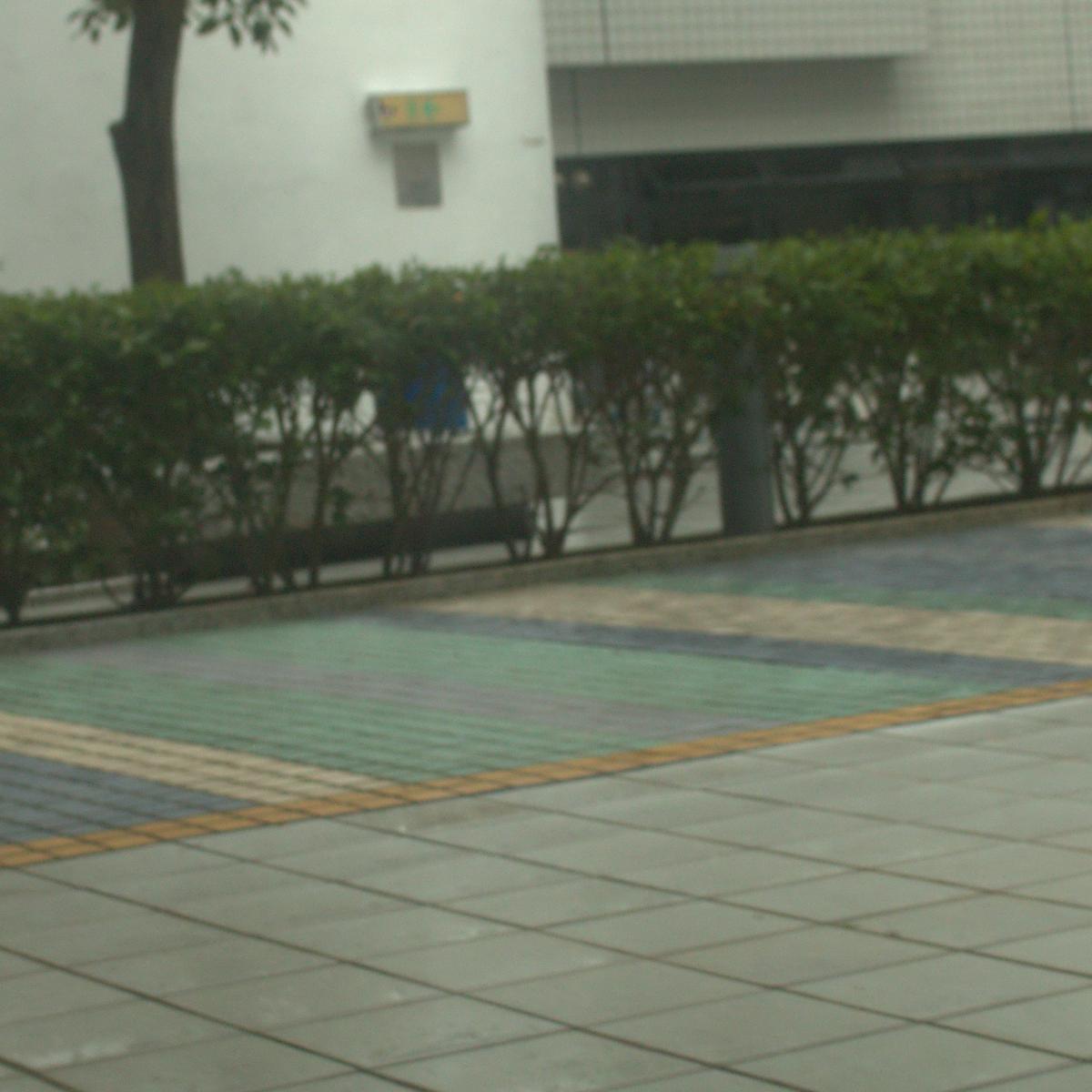}&
\includegraphics[width=0.13\linewidth]{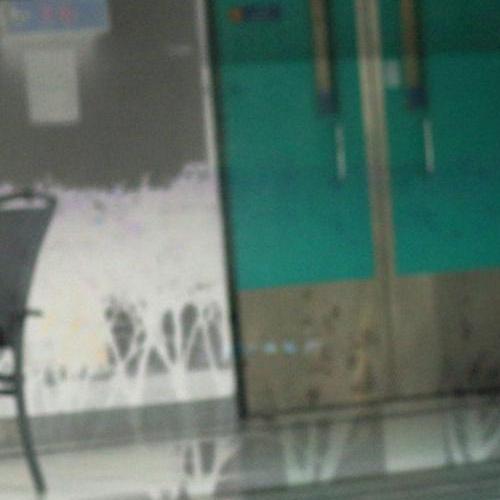}&
\includegraphics[width=0.13\linewidth]{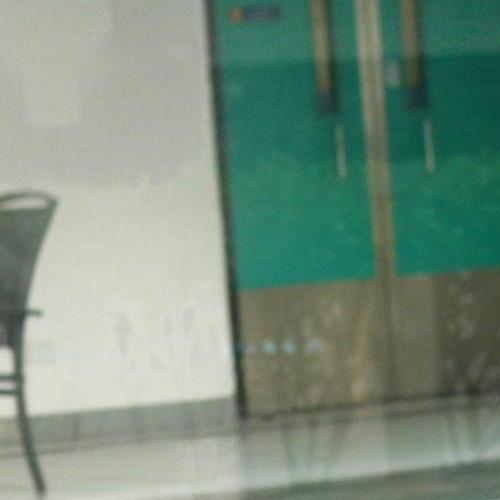}&
\includegraphics[width=0.13\linewidth]{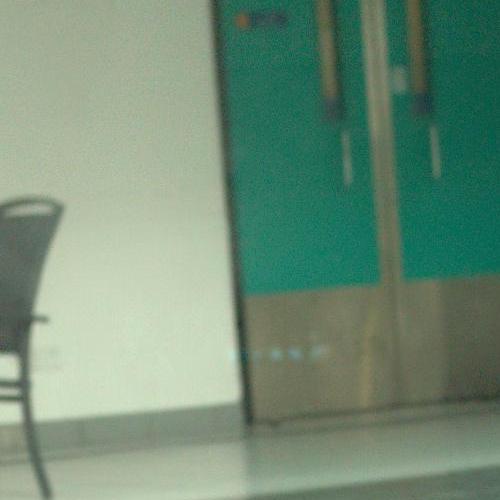}&
\includegraphics[width=0.13\linewidth]{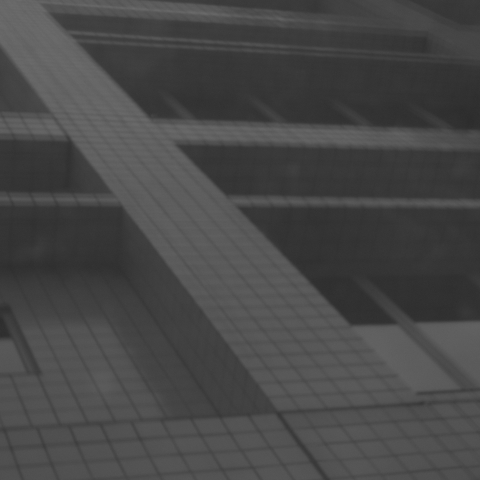}&
\includegraphics[width=0.13\linewidth]{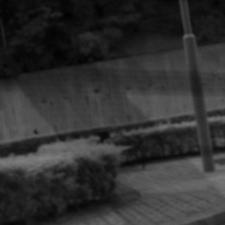}&
\includegraphics[width=0.13\linewidth]{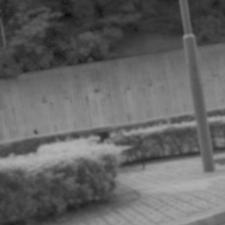}\\

&RGB R & Closeup & Closeup & Closeup & Pol R & Closeup & Closeup\\
\end{tabular}
%\vspace{1mm}
\caption{The visualization of M-R in different data spaces. If $M-R$ is applied other than raw data space, undesirable residuals will appear. For RGB data, we get 3 types of M-R: (1) ``ISP M-R'' means do $M-R$ on images after ISP. (2) ``Gamma M-R'' means to use $M^{2.2}-R^{2.2}$ to simulate gamma decompression for $M$ and $R$, which is a common way used in previous methods. (3) ``Raw M-R'': do $M-R$ on raw data. For the gray-scale polarization data, we use gamma correction to simulate the ISP, compared with directly on raw.}
\label{fig:MMR_sample}
\end{figure*}

\section{Related Work}

\paragraph{Single image reflection removal.}
Most single image reflection removal methods~\cite{fan2017generic, zhang2018single, eccv18refrmv_BDN, Yang_2019_CVPR} rely on various assumptions. Considering image gradients, Arvanitopoulos et al.~\cite{Arvanitopoulos_2017_CVPR} propose the idea of suppressing the reflection, and Yang et al.~\cite{Yang_2019_CVPR} propose a faster method based on convex optimization. These methods fail to remove sharp reflection. Under the assumption that transmission is always in focus, Punnappurath et al.~\cite{Punnappurath_2019_CVPR} design a method based on dual-pixel camera input. 
For most deep learning based approaches, training data is critical for good performance. CEILNet~\cite{fan2017generic}, Zhang et al.~\cite{zhang2018single} and BDN~\cite{eccv18refrmv_BDN} assume reflection is out of focus and synthesize images to train their neural networks. CEILNet~\cite{fan2017generic} estimates target edges first and uses it as guidance to predict the transmission layer. Zhang et al.~\cite{zhang2018single} use perceptual and adversarial losses to capture the difference between reflection and transmission. BDN~\cite{eccv18refrmv_BDN} estimates the reflection image, which is then used to estimate the transmission layer. These methods~\cite{zhang2018single,fan2017generic,eccv18refrmv_BDN} work well when reflection is more defocused than transmission but fail otherwise. 
To break the limitation of using solely synthetic data, Zhang et al.~\cite{zhang2018single} and Wei et al.~\cite{wei2019single_ERR} collect real-world datasets for training. However, their datasets have misalignment issues and do not contain sufficient diversity. Wei et al.~\cite{wei2019single_ERR} propose to use high-level features that are less sensitive to small misalignment to calculate losses.
 To obtain more realistic and diverse data, 
Wen et al.~\cite{Wen_2019_CVPR_Linear} and Ma et al.~\cite{Ma_2019_ICCV} propose methods to synthesize data using a deep neural network and achieve better performance and generalization. Though the data is more perceptually appealing, physical authenticity remains a doubt.
%These two approaches get more realistic data and improve performance and generalization. However, their synthesis data rely on training data and method. 

\paragraph{Polarization-based reflection removal.}
Polarization is known to be useful in image reflection removal since decades ago~\cite{Schechner1999PolarizationbasedDO,Fraid1999}.
Schechner et al. \cite{Schechner1999PolarizationbasedDO} and Bronstein et al.~\cite{bronstein2005sparse} utilize independent component analysis to separate reflection and transmission images. With the assumption of unpolarized light sources, Kong et al.~\cite{kong14pami} proposed an optimization method to automatically find the optimal separation of the reflection and transmission layer. Wieschollek et al.~\cite{eccv2018/Wieschollek} combine deep learning with a polarization-based reflection removal method. Different from previous works, they eliminate a number of assumptions (e.g., the glass must be perfectly flat) and propose a pipeline to synthesize data with polarization information from regular RGB images. However, all light sources are still assumed to be unpolarized.

\paragraph{Multi-image reflection removal}
Polarization-based reflection removal methods are a special category of multi-image approaches. Agrawal et.al~\cite{DBLP:journals/tog/AgrawalRNL05} use a pair of flash/no-flash images. Many works~\cite{szeliski2000layer,Sarel2005,sarel2004separating, li2013exploiting,guo2014robust,han2017reflection,xue2015computational} move the camera to exploit the relative motion between reflection and transmission for reflection removal, while most works assume that motion of the reflection layer is larger than that of the transmission layer. Sarel and Irani~\cite{sarel2004separating, Sarel2005} assume that both reflection and transmission should be static. Li et al.~\cite{li2013exploiting} use SIFT-flow to align the images to make a pixel-wise comparison under the assumption that the background dominates in the mixed image. Xue et al.~\cite{xue2015computational} also require that objects in reflection and transmission are roughly static. Han et al.~\cite{han2017reflection} require the transmission to be more dominant than the reflected scenes.

\begin{center}
\begin{table*}%[t]
\centering
\renewcommand{\arraystretch}{1.2}
\begin{tabular}{lccccccc}
\Xhline{1.0pt}
                & Glass type  & Data format & Scene & Alignment & Intensity decay & Raw \\ \hline
Zhang et al.~\cite{zhang2018single} & 1  & $M, B$  & 110 & Misalignment (calibrated) & Yes& No     \\ 

Wei et al.  ~\cite{wei2019single_ERR}& 1 & $M, B$  & 450 & Misalignment & Yes& No      \\ 
SIR benchmark~\cite{wan2017benchmarking}& 3 & $M, R, B$ & 100+20+20 & Misalignment (calibrated) & Yes& No     \\ 
Ours    & $>$100  & $M, R, T$  & 807  & Perfect Alignment & Small& Yes    \\ \Xhline{1.0pt}
\end{tabular}
\vspace{1mm}
\caption{Comparison between our collected dataset and others. Our dataset has more diverse glass types, perfect alignment, and little intensity decay. Besides, since we provide raw data, we can synthesize new $\{M, R, T\}$ triples faithful to the real image formation process.}
\label{table:Dataset comparison}
\end{table*}
\end{center}

\section{M-R Dataset}
Real-world reflection removal datasets~\cite{zhang2018single,wei2019single_ERR} are limited in quantity and diversity because of the complicated data collection procedure and the difficulty of acquiring ground-truth reflection and transmission. We propose a new method named M-R to collect paired data for reflection removal. A triple $\{M, R, T\}$ is collected for each scene where $M, R, T$ are the mixed image, the reflection image, and the transmission image, respectively. 
 
\paragraph{Polarization information}
We use the PHX050S-P polarization camera, which is equipped with an IMX250MZR CMOS. This sensor captures an image with four different polarizer angles in one single shot. Each polarization pixel consists of $2 \times 2$ units with four sub-pixels corresponding to the polarization angles $0^{\circ}, 45^{\circ}, 90^{\circ}, 135^{\circ}$. The light intensity passing through a polarizer follows Malus' law~\cite{hecht2002optics}:
\begin{align}
    I_{out} = I_{in}cos^2(|\phi-\theta|),
\end{align}
where $\theta$ is the angle of polarizer, and $\phi$ is the polarization angle of incoming light. Note that the equations related to polarization hold only for raw data that is linear to light intensity, and thus we adopt the RAW format in our dataset.
The resolution of each captured RAW image is $2048\times2448$. We extract sub-pixels with the same polarization angle to form an image, and we can get four images with resolution $1024\times1224$. The value range of each pixel is from 0 to 4095. Let $I$ be the light intensity, and let $I_{np}, I_p$ be the intensity for unpolarized light and linear polarized light. The degree of polarization $\rho$ equals to $I_p/(I_p+I_{np})$. Then we define $I_1,I_2,I_3,I_4$ as the light intensity passed through 4 angles $0^{\circ}, 45^{\circ}, 90^{\circ}, 135^{\circ}$. According to the properties of polarization, we have: 
\begin{align}
\label{eq:pol}
&I = (I_1 + I_3 + I_2 + I_4)/2, \\
\label{eq:pol2}
&\textbf{$\rho$} = \frac{\sqrt{((I_1 - I_3)^{2} + (I_2 - I_4)^{2})}}{I}, \\
\label{eq:pol3}
&\textbf{$\phi$} = \frac{1}{2}arctan\frac{I_2-I_4}{I_1-I_3}.
\end{align}

\begin{figure}[h]
\centering
\begin{tabular}{@{}c@{\hspace{1mm}}c@{\hspace{1mm}}c@{}}
\includegraphics[width=0.32\linewidth]{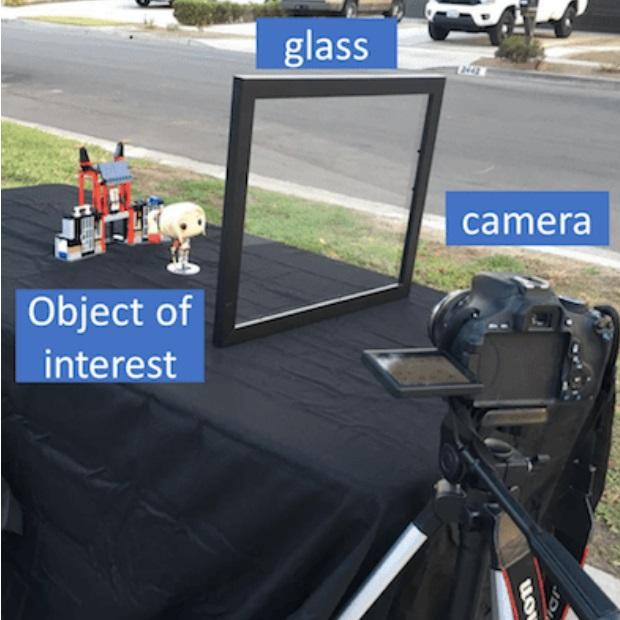}&
\includegraphics[width=0.32\linewidth]{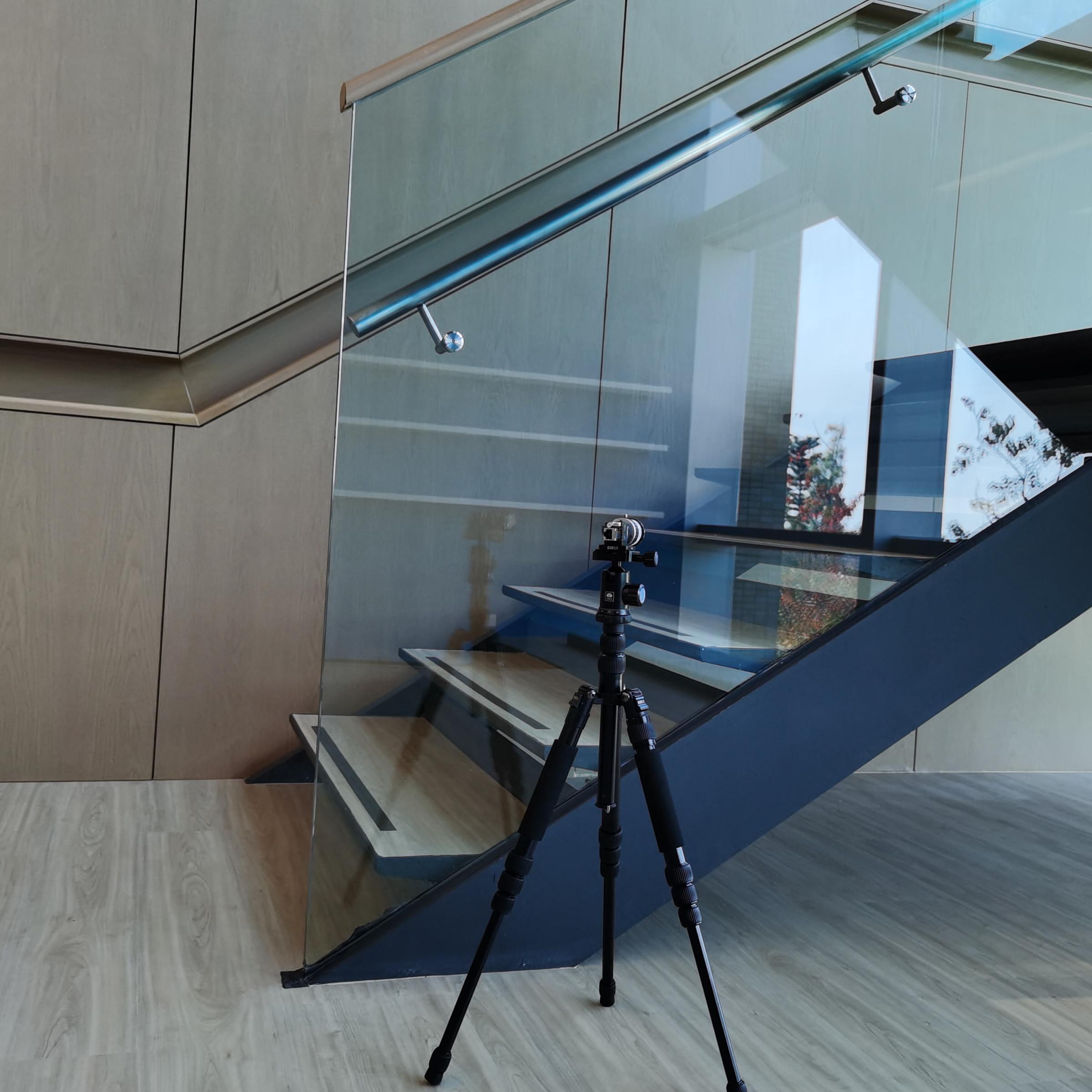}&
\includegraphics[width=0.32\linewidth]{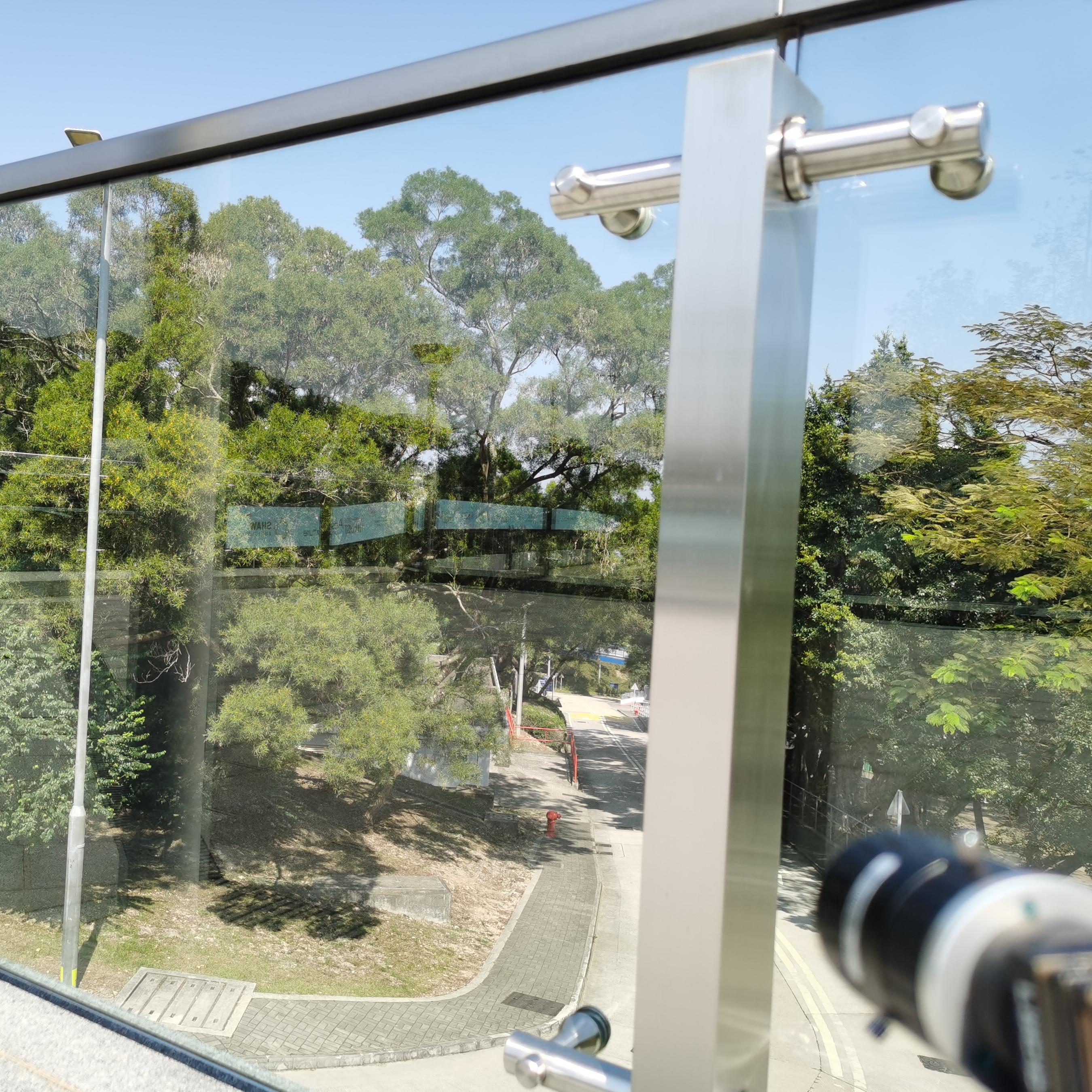}\\
Zhang et al.~\cite{zhang2018single} & Our $M$& Our $M$\\
\includegraphics[width=0.32\linewidth]{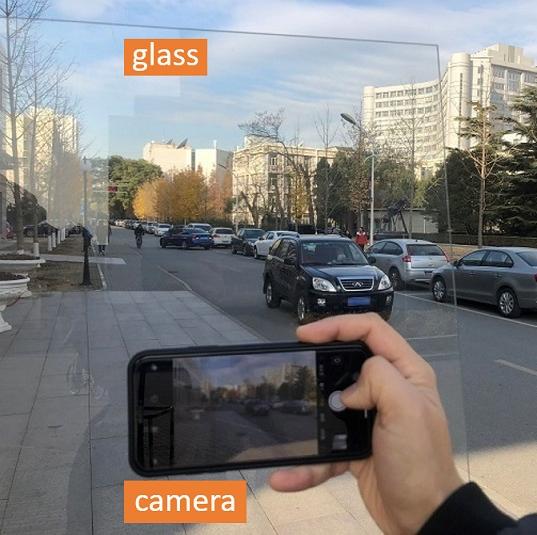}&
\includegraphics[width=0.32\linewidth]{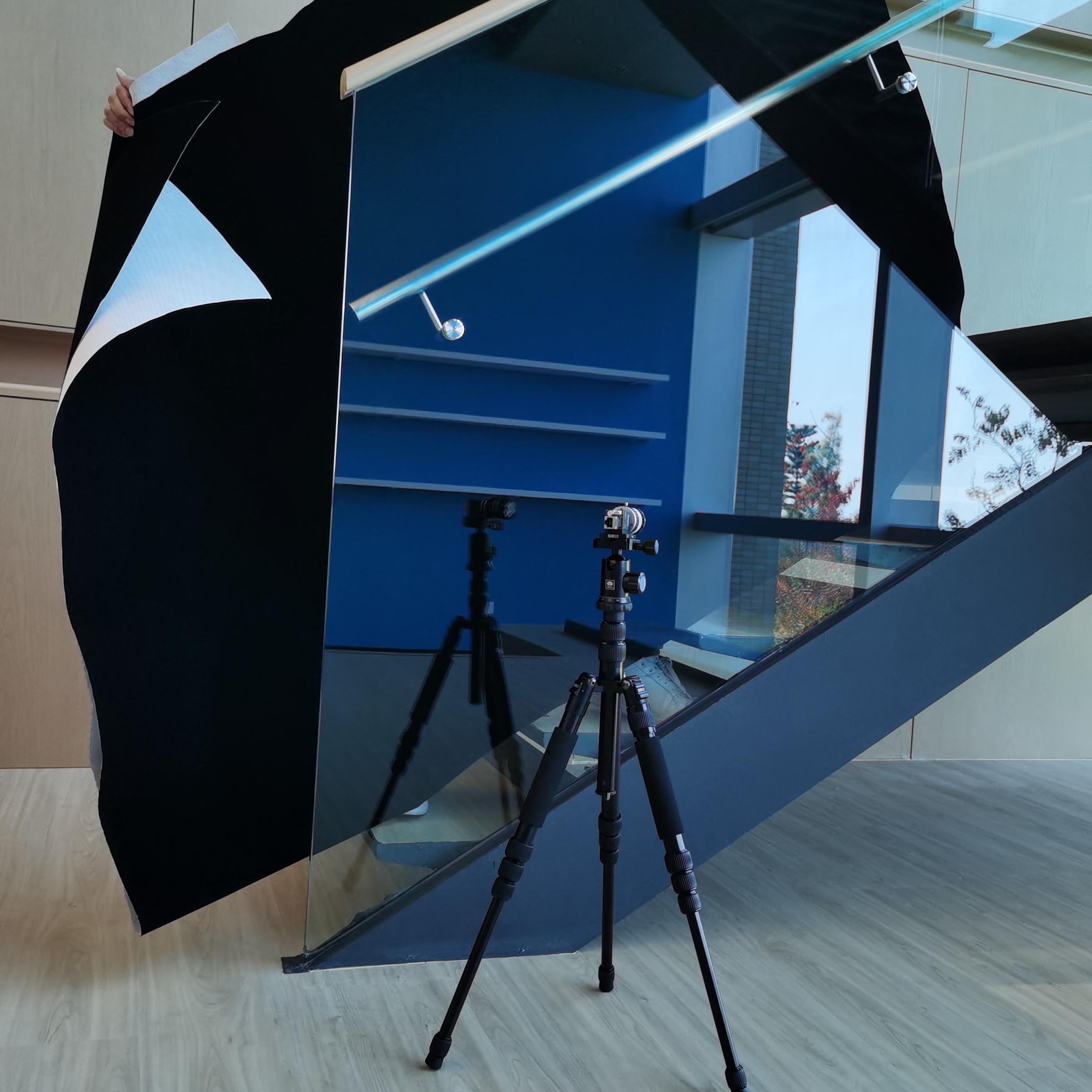}&
\includegraphics[width=0.32\linewidth]{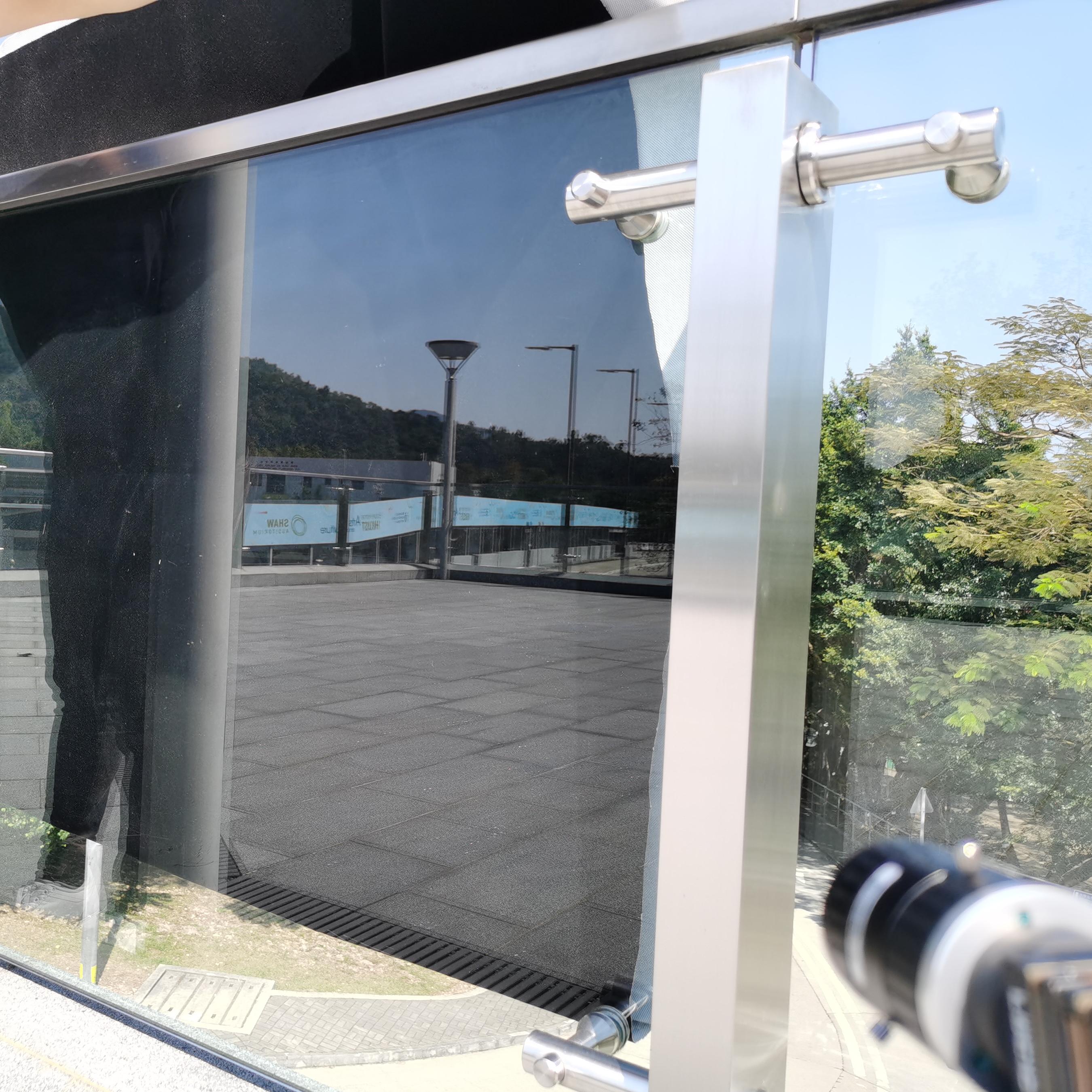}\\
Wei et al.~\cite{wei2019single_ERR} & Our $R$& Our $R$\\

\end{tabular}
%\vspace{1mm}
\caption{Comparison between our data collection pipeline and prior methods. Previous methods use removable, relatively thin, and non-colored glass to avoid misalignment and color distortion. Since we do not enforce these constraints, we can utilize a diverse set of glass types that exist in our daily life. }
\label{fig:PriorData}
\end{figure}

\paragraph{Data collection pipeline}
Fig. \ref{fig:PriorData} shows the comparison between our pipeline and previous~\cite{zhang2018single, wei2019single_ERR, wan2017benchmarking}. Previous methods take a photo in front of glass as a mixed image $M$ and then remove the glass to take another one as the transmission $T$ so that the difference between background $B$ and transmission $T$ is ignored. As mentioned before, $M$ is the sum of $R$ and $T$ (not $B+R$). Therefore, inferring $T$ is believed to be easier than $B$. However, it is relatively difficult to capture $T$ directly because all the reflection must be blocked. Therefore, we capture $M$ and $R$ only and then obtain $T=M-R$.

While prior work~\cite{Wen_2019_CVPR_Linear, Ma_2019_ICCV} claims that the combination of reflection and transmission is beyond linearity, we argue that the non-linearity is introduced by ISP pipeline when operating in RGB space.
On the other hand, there is no such problem for raw data since the voltage on the sensor is linearly correlated with the intensity of light. Therefore, Eq. (\ref{eq:MTR}) holds, and we can obtain a transmission image $T$ directly by $M - R$. Fig. \ref{fig:MMR_sample} shows the difference between RGB data and raw data. It is clear that our formulation conforms with reality, and the direct subtraction removes reflection perfectly. To the best of our knowledge, we are the first to use $M-R$ as ground truth on raw data.

To ensure perfect alignment between $M$ and $R$, we use a tripod to fix the camera and take the polarized images remotely controlled by a computer. We first use a piece of black cloth to cover the back of the glass to block all transmission $T$ to obtain a clear reflection $R$. Then we remove the cloth to collect the mixed image $M$. To ensure the intensity of reflection are the same in $M$ and $R$, we set the camera to manual model with relatively long exposure time to avoid noise.

\begin{figure*}[t!]
\centering
\includegraphics[width=1.0\linewidth]{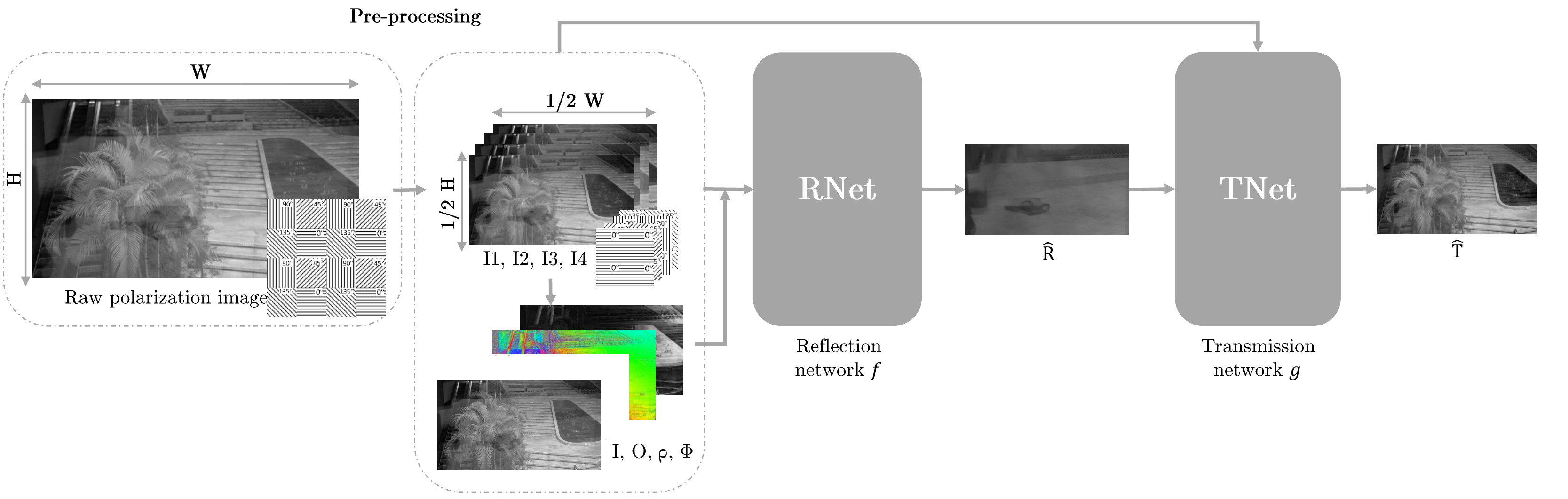}
\vspace{-3mm}
\caption{The overall architecture of our model. The raw input goes through a simple pre-process before the network. The reflection network $f$ is designed to get $f(M)=\hat R$. The refinement network $g$ estimates $\hat T$ based on $\hat R$ and $M$, where $ g(M,\hat R) = \hat T$. }
\label{fig:Framework}
\end{figure*}

\paragraph{Analysis of M-R} Table \ref{table:Dataset comparison} shows the comparison between our dataset and previous datasets. Compared with previous methods, M-R has the following advantages:

     a). \textbf{More diversity}. Previous methods require the glass to be thin, non-colored, removable, and flat. As long as the transmission is clear, we do not make such assumptions on the glass. Therefore, it is possible for us to utilize numerous glass in our daily life, such as glass doors and windows. The glass can be flat or curved, thin or thick, colored, or non-colored. We are even able to record dynamic scenes if the reflection is static.

     b). \textbf{Simplified task.}
     Since $B$ might be different with $T$ in color, intensity, and position, using $B$ as ground truth introduces extra problems in reflection removal. Estimating $T$ is an equally useful and simplified task. Our dataset has provided perfectly aligned pairing data.
     
     c). \textbf{Improved simulation.}
     Even if we use our method, collecting paired data is time-consuming. Since previous methods have the misalignment problem, they can not correctly obtain $R$ by $M-B$. Besides, they use RGB images instead of raw images, so non-linearity in intensity is introduced. Derived from the linearity discussed above, we can use $M = a*R + b*T$ directly to simulate various realistic data where $a$ and $b$ varies from $0$ to $1$ with unpaired $R$ and $T$.

\paragraph{Data cleaning} 
To improve the quality of our dataset, we calculate the mean intensity ratio for each pair of $R$ and $T$, and discard the pairs if the ratio is greater than 10 or smaller than 0.1. As in this situation, either $R$ or $T$ is perceptually invisible. Negative values after subtraction, due to noise, are set to zero. If there is more than one layer of glass, we crop the image to keep only the part with a single layer. Polarization can be calculated correctly only if each polarization image is correct. Hence, we need to pay special attention to overexposed areas. We calculate an overexposure mask $O$ based on the intensity of $I_1,I_2,I_3,I_4$.

\begin{align}
O(x)=
\begin{cases}
0,& \text{max}\{I_1(x),I_2(x),I_3(x),I_4(x)\} > \delta    \\
1,& otherwise
\end{cases}
\label{eq:Overexp}
\end{align}
where $\delta$ is a threshold and we use $\delta=0.98$ here.

\section{Method}

\subsection{Reflection-Based Framework}
%Resolution

\begin{figure}[ht]
\centering
\includegraphics[width=0.8\linewidth]{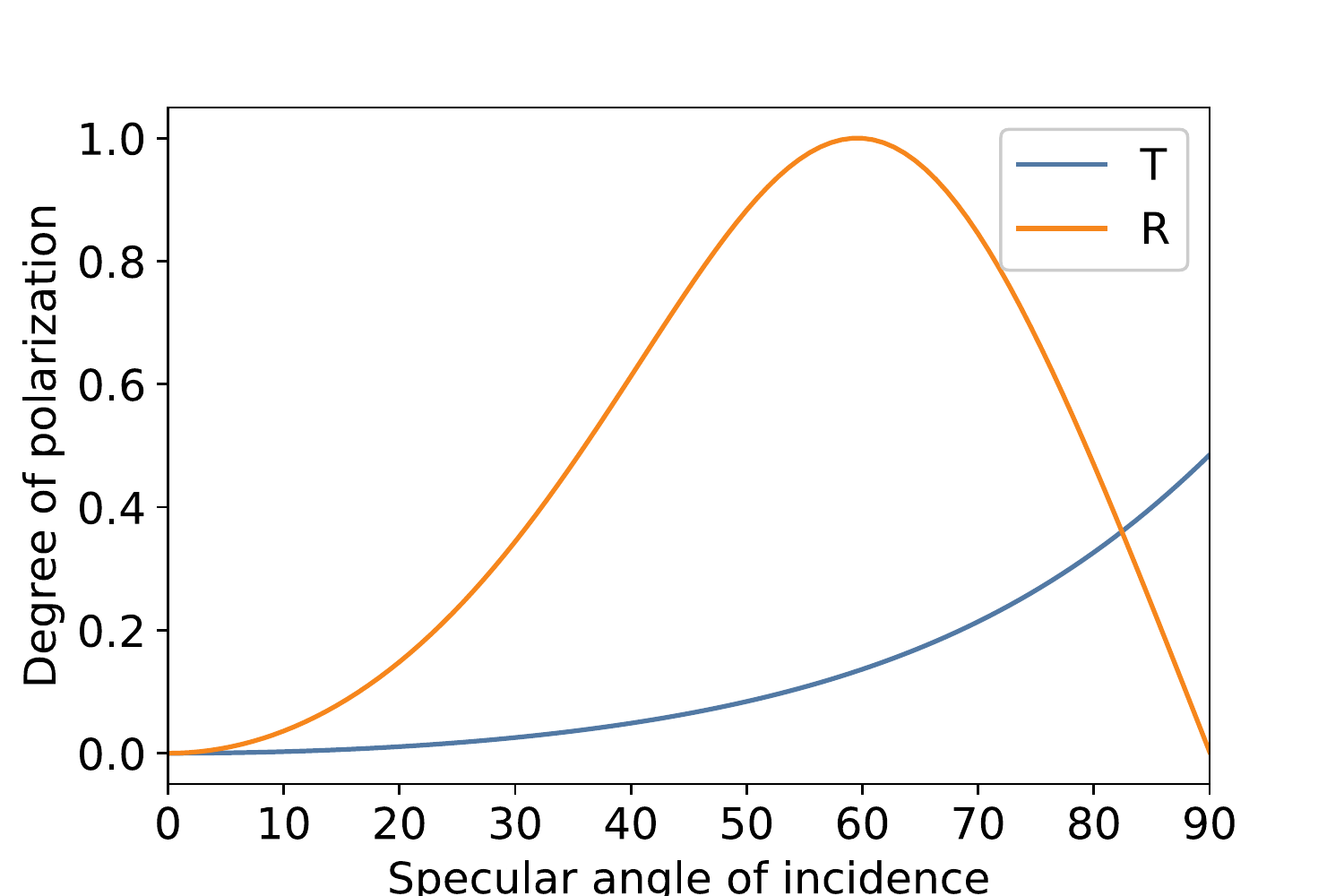}
%\vspace{1mm}
\caption{
% For specular reflection, the degree of polarization $\rho$ for transmission light and reflected light. We could see that the $\rho$ of $R$ is more significant than $T$ in most areas. This curve aims for a dielectric (n=1.7)
The degree of polarization $\rho$ for transmitted light $T$ and reflected light $R$ for a specific dielectric (n=1.7). For most incident angles, $\rho$ for $R$ is more significant than $T$.
}
\label{fig:Fresnel}
\end{figure}

Unpolarized light reflected from the glass surface or passed through the glass becomes partially polarized. The degree of polarization, $\rho$, depends on the property of glass and the angle of incidence. For a specific type of glass with refractive index $n = 1.7$, Fig. \ref{fig:Fresnel} shows how the degree of polarization changes. Based on this fact, Kong et al.~\cite{kong14pami} and Wieschollek~\cite{eccv2018/Wieschollek} propose two methods for reflection removal. However, in the real world, unpolarized light sources assumption doesn't hold well because partially polarized light sources exist commonly, and reflection exists not only through glass surfaces. These methods would then fail~\cite{eccv2018/Wieschollek, kong14pami}. Different from Wieschollek et al.~\cite{eccv2018/Wieschollek} and Kong et al.~\cite{kong14pami}, we do not assume all light sources are unpolarized. We utilize the fact that the $\rho$ of transmission is quite different. Hence we propose to use a deep learning based and two-stage method to catch the differences between reflection and transmission and separate them.

Fig. \ref{fig:Framework} shows an overview of our framework. Our method takes a multi-channel image as input. The first 4 channels, $I_1, I_2, I_3, I_4$ are extracted from mixed image $M$ for each polarization angle. The next 4 channels are $I, \rho, \phi, O$, calculated from Eq. \ref{eq:pol}, \ref{eq:pol2}, \ref{eq:pol3}, \ref{eq:Overexp}. The final network output is a one-channel image, the recovered transmission $\hat{T}$, the same size as $I_1$, that is half of $M$ in width and height.

% Explain two stages. What is? Why?
There are two stages in our process. The first stage is dedicated to estimating reflection $\hat R$, and the second is for transmission $\hat T$ with estimated $\hat R$. We use a two-stage design for two reasons. Firstly, reflection contributes a lot to mix image and has a strict relationship on RAW space (Eq. \ref{eq:MTR}). Furthermore, as discussed above, reflection and transmission are quite different in terms of polarization. The separated decoders for them are helpful to learn specific features. BDN~\cite{eccv18refrmv_BDN} also observes the importance of reflection and improves performance by training a bidirectional network. However, their performance relies on an assumption to make $R$ and $T$ more different: the reflection is blurry. Undoubtedly, their model cannot distinguish $R$ and $T$ well when reflection is sharp. Note that if without polarization, such design may deteriorate the performance as the difference between them becomes subtle in regular image data.

% Express important of reflection
% Most previous works ignore the importance of estimating reflection. However, Yang et al~\cite{eccv18refrmv_BDN} observe it and improve the performance by training a bidirectional network. They rely on the assumption that reflection is blurry to differentiate?????.
%We design a reflection-based framework based on the difference of polarization information between reflection and transmission. Note this framework works when difference exists, otherwise it might even degrade the performance. In total, we have $I_1,I_2,I_3,I_4$, $I$, $\phi$, $\rho$ and $O$ as the input for our network. We use a reflection network $f$ and transmission network $g$ to estimate the transmission. We also tried using a single network to estimate $R$ and $T$ together; however, the results are not quite good. Since the property of reflection and transmission has a big difference, we believe it's better to use two networks for them.

%We use the reflection network $f$ to estimate $R$ directly based on $M$. After extract 4 images at different polarization angles and calculate the related polarization information. We concatenate them together to our reflection network $f$. 

\subsection{Loss function}
\paragraph{PNCC loss} 

In general, reflection and transmission images would be different on most pixels. We propose a perceptual normalized cross-correlation (PNCC) loss to minimize the correlation between estimated reflection and transmission on different feature maps. Our PNCC loss is defined on different feature maps of VGG-19 ~\cite{simonyan2014very}. Given two images $I_A$ and $I_B$, we try to calculate the NCC of their feature maps. In practice, the monotonicity is not right in extreme cases where the intensity between $R$ and $T$ has a big difference. Therefore, we normalize $I_A, I_B$ to $[0,1]$, denoted as $\tilde I_A, \tilde I_B$. The PNCC loss is defined as follows:
\begin{align}
    L_{PNCC}(I_A, I_B)=
    &\sum_{l=1}^n{NCC(v_l(\tilde I_A), v_l(\tilde I_B))
    },
\end{align}
where $v_l$ denotes the $l$-th layer feature maps of VGG-19~\cite{simonyan2014very}. In practice, we use three layers 'conv2\_2','conv3\_2','conv4\_2'. PNCC can also be applied using another pre-trained neural network.

\begin{figure}
\centering
\includegraphics[width=0.8\linewidth]{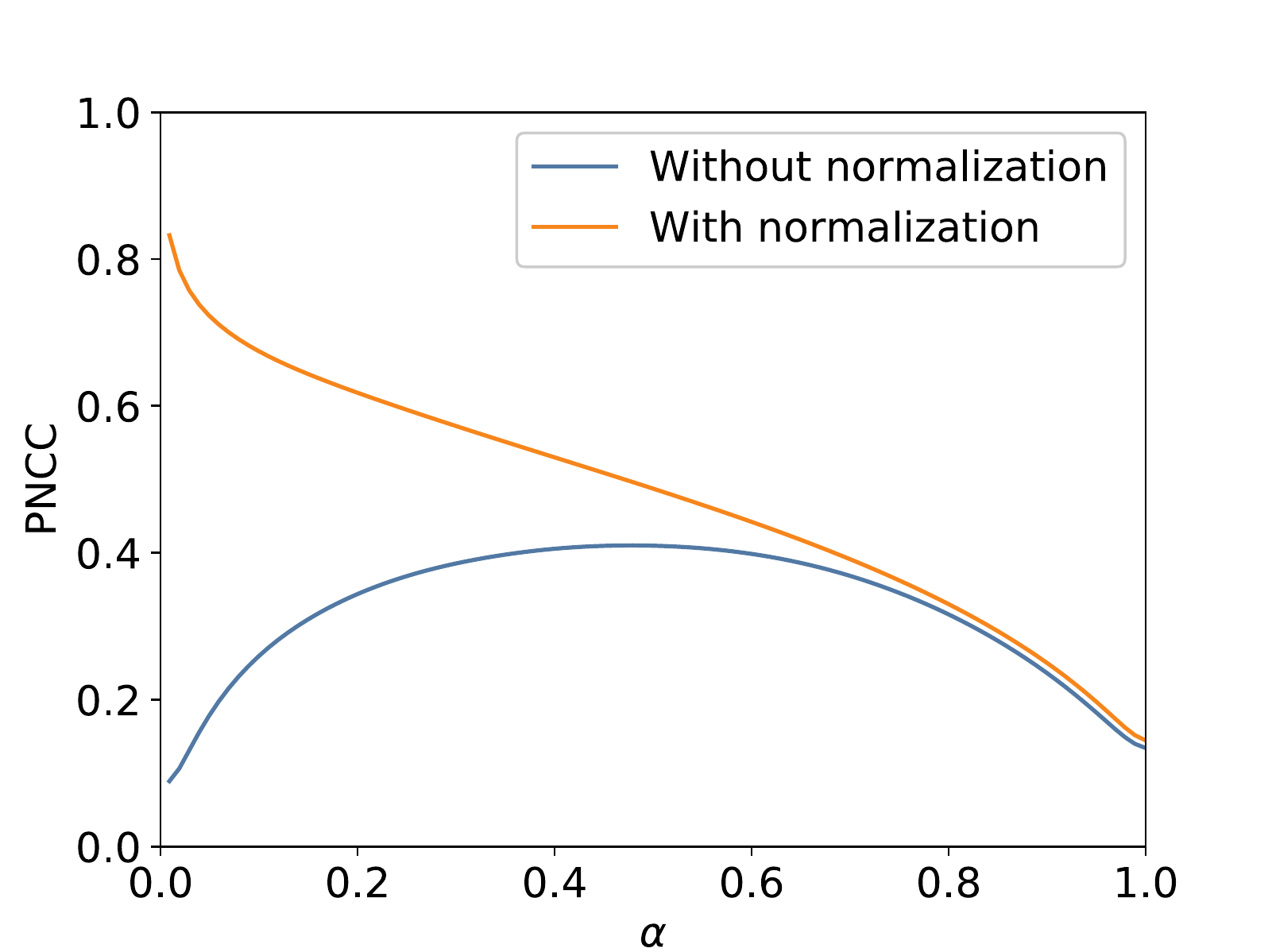}
% %\vspace{1mm}
% \caption{Analysis of the monotonicity of the PNCC loss.}
%\caption{Monotonicity of the PNCC loss. The proposed PNCC monotonically decreases as the input pair gets mix more. In the meantime, without normalization to the input, monotonicity is not preserved.}
\caption{The monotonicity of the PNCC loss. The proposed PNCC monotonically decreases as the input pair gets mix more after applying normalization to the input.}

\label{fig:PNCC-monotonicity}
\end{figure}

Fig. \ref{fig:PNCC-monotonicity} shows the monotonicity of PNCC and the impact of normalization. We choose 100 pairs of images randomly from the dataset used in~\cite{zhang2018single}. For each pair $(R, T)$ we generate synthetic data $I_A$ and $I_B$ by:
\begin{align}
I_A=T+(1-\alpha)*R, I_B=\alpha*R,
\end{align}
where $\alpha$ is sampled from 0.01 to 1. When $\alpha = 1$, $I_A$ and $I_B$ are completely two different images, PNCC is the lowest. When $\alpha = 0.01$, $I_A$ contains most part of $I_B$, PNCC is the largest, but the non-normalized version is not. Our PNCC loss can also be applied to other image decomposition tasks. More results are demonstrated in experiments.

\paragraph{Perceptual loss}
The perceptual loss~\cite{Johnson2016Perceptual} has been proved effective on various computer vision tasks~\cite{Lei_2019_CVPR,zhang2018single,Chen2017}. In our task, we modify it to account for the overexposed area. Given the overexposure mask $O$, the perceptual loss is defined as:
\begin{align}
    &L_{p}(T,\hat T)=\sum_{l=1}^n{\beta_l |v_l(O*T) -  v_l(O*\hat T)|_1}.
\end{align}
$\beta_l$ is the weight for the $l$-th layer. Following Chen et al.~\cite{Chen2017}, we initialize $\beta_l$ based on the number of parameters in each layer and we adopt 6 layers 'conv1\_1', 'conv1\_2', 'conv2\_2', 'conv3\_2', 'conv4\_2', and 'conv5\_2'. 

In total, the loss function we optimize is the sum of PNCC loss between $\hat R$ and $\hat T $ and perceptual loss.

\begin{figure*}[h]
\centering
\begin{tabular}{@{}c@{\hspace{1mm}}c@{\hspace{1mm}}c@{\hspace{1mm}}c@{\hspace{1mm}}c@{\hspace{1mm}}c@{\hspace{1mm}}c@{}}
\rotatebox{90}{\small \hspace{7mm} Blur }&
\includegraphics[width=0.155\linewidth]{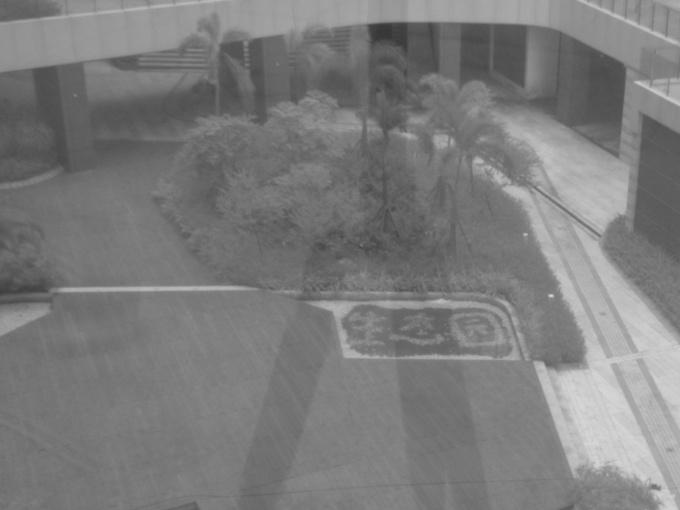}&
\includegraphics[width=0.155\linewidth]{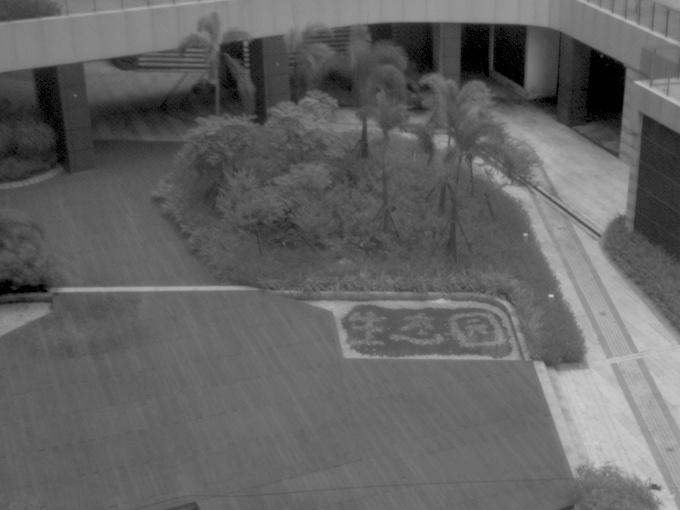}&
\includegraphics[width=0.155\linewidth]{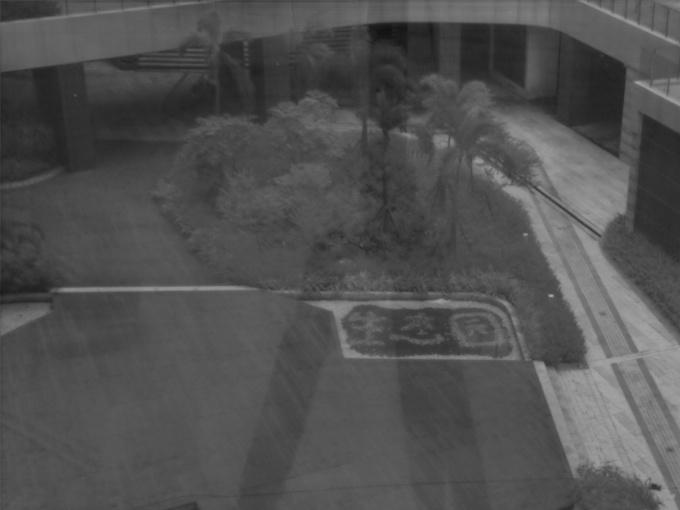}&
\includegraphics[width=0.155\linewidth]{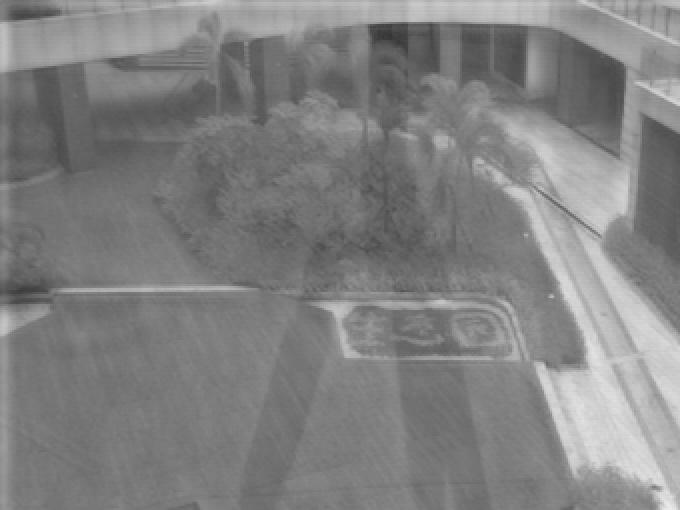}&
\includegraphics[width=0.155\linewidth]{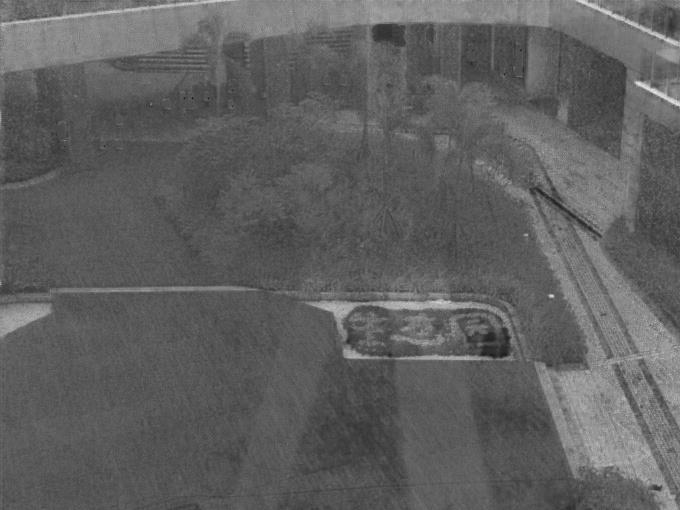}&
\includegraphics[width=0.155\linewidth]{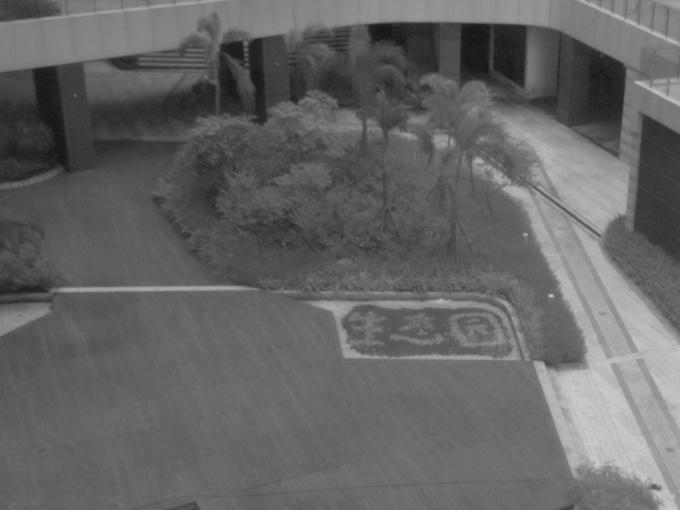}\\
\rotatebox{90}{\small \hspace{3mm} Ghost Cues}&
\includegraphics[width=0.155\linewidth]{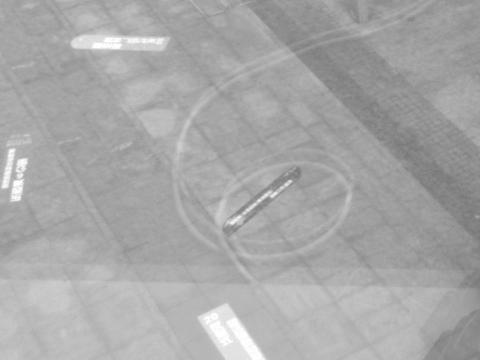}&
\includegraphics[width=0.155\linewidth]{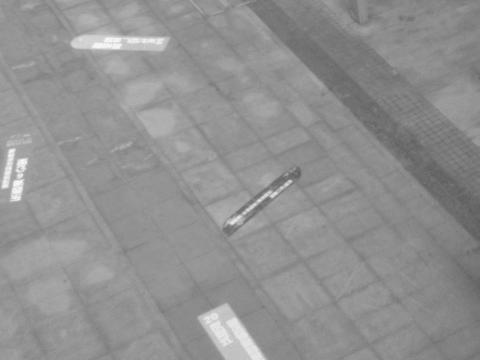}&
\includegraphics[width=0.155\linewidth]{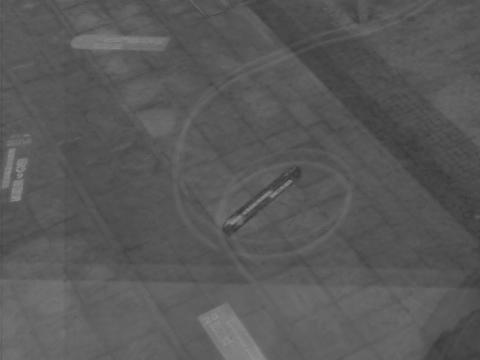}&
\includegraphics[width=0.155\linewidth]{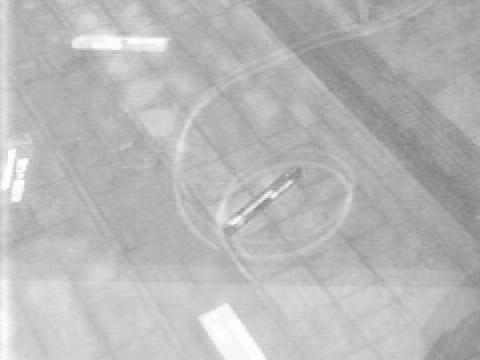}&
\includegraphics[width=0.155\linewidth]{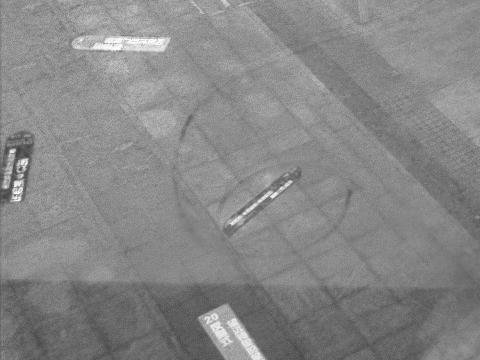}&
\includegraphics[width=0.155\linewidth]{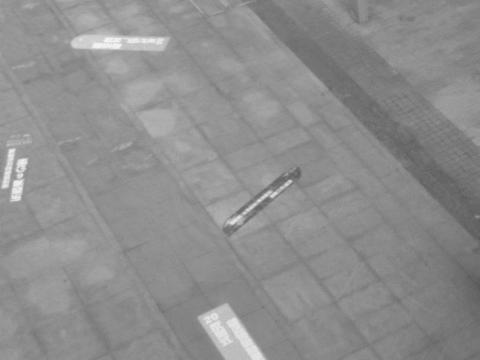}\\
\rotatebox{90}{\small \hspace{7mm} Sharp }&
\includegraphics[width=0.155\linewidth]{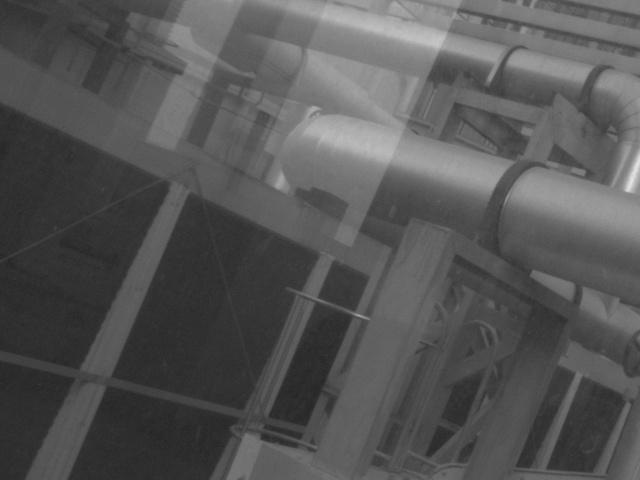}&
\includegraphics[width=0.155\linewidth]{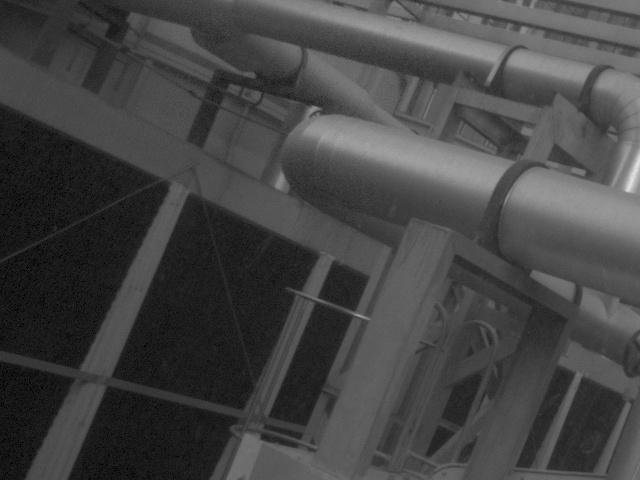}&
\includegraphics[width=0.155\linewidth]{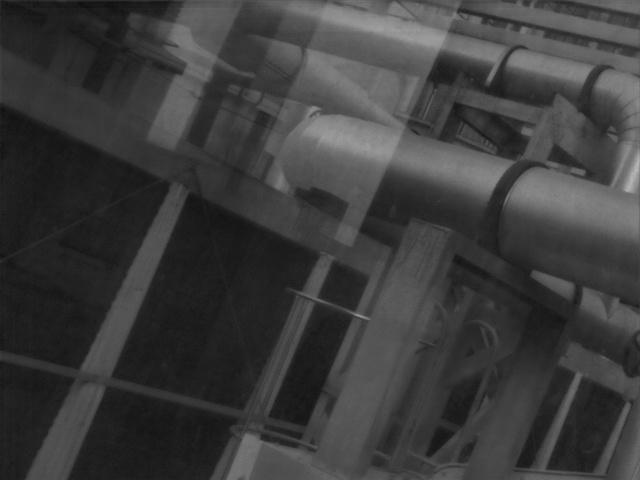}&
\includegraphics[width=0.155\linewidth]{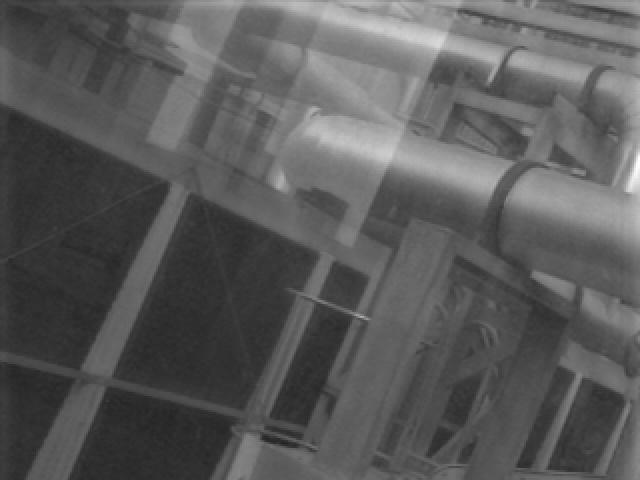}&
\includegraphics[width=0.155\linewidth]{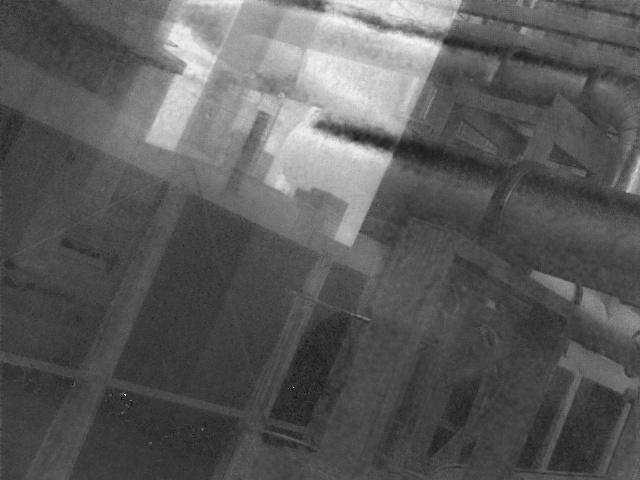}&
\includegraphics[width=0.155\linewidth]{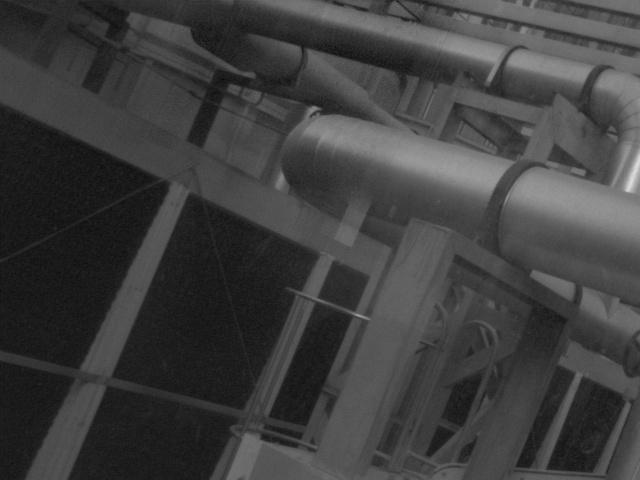}\\
&Input & Ground Truth & Zhang et al.~\cite{zhang2018single} & Wei et al.~\cite{wei2019single_ERR} &  \footnotesize{Wieschollek et al.}
\cite{eccv2018/Wieschollek} & Ours\\
\end{tabular}
%\vspace{1mm}
\caption{Perceptual comparison between our method and others. Our method is able to handle different types of reflection.}
\label{fig:Perceptual Comparison}
\end{figure*}

\subsection{Implementation}

To improve the performance of our model, we augment the input to the network with the hypercolumn features extracted from the VGG-19 network~\cite{simonyan2014very}. In particular, we extract 'conv1\_2' from the VGG-19 network for $I_1, I_2, I_3, I_4, I$ and upsample the layers bilinearly to match the resolution of the input image.  Since our data is in RAW format and pre-trained VGG-19~\cite{simonyan2014very} was trained on ImageNet dataset~\cite{Deng2009} in RGB space, we first apply a gamma correction to the raw input and then feed them into the network. We adopt U-Net~\cite{Ronneberger2015} as our network architecture for both $f$ and $g$. We modify the kernel size of the first layer to $1\times1$ and use it to reduce the dimensionality of the augmented input~\cite{Chen2017}. At the training, we first train $f$ and $g$ together for 200 epochs using Adam optimizer and learning rate 0.0001. Then we decay the learning rate to 0.00001 and train for 50 more epochs.

\section{Experiments}
\subsection{Experimental procedure}
%What hypotheses are you testing?

%   baseline to adopt
\paragraph{Baselines.} 
We compare our method with several state-of-the-art reflection removal approaches, including both deep learning and traditional methods.
Specifically, in the deep learning track, we choose Zhang et al.~\cite{zhang2018single}, Wei et al.~\cite{wei2019single_ERR}, BDN~\cite{eccv18refrmv_BDN}, Wieschollek et al.~\cite{eccv2018/Wieschollek}, and Wen et al.~\cite{Wen_2019_CVPR_Linear}.
For fairness, we re-train models on our M-R dataset using official source codes for Zhang et al.~\cite{zhang2018single} and Wei et al.~\cite{wei2019single_ERR}. For BDN~\cite{eccv18refrmv_BDN} and Wieschollek et al.~\cite{eccv2018/Wieschollek}, we directly use the available pre-trained models since no training codes are available. For Wen et al.~\cite{Wen_2019_CVPR_Linear}, as their training requires additional alpha matting masks that are not available in our task, we also use their pre-trained model.

For polarization based methods, we choose Kong et al.~\cite{kong14pami}, Schechner et al.~\cite{2000Schechner} and Fraid et al.~\cite{Fraid1999}. Third-party implementations by Wieschollek et al.~\cite{eccv2018/Wieschollek} are used. We also evaluate the convex optimization based method by Yang et al.~\cite{Yang_2019_CVPR} using their official source codes.

DoubleDIP~\cite{DoubleDIP} is an unsupervised image decomposition model, but it fails in our setting. The possible reason is that DoubleDIP holds a simple assumption that a mixed image is composed of two images with spatial-invariant coefficients, but real-world data break the assumption.

\begin{table}[]
\small
\centering
\renewcommand{\arraystretch}{1.2}
\begin{tabular}{lcccc}
\Xhline{1.0pt}
& \multicolumn{2}{c}{Transmission} & \multicolumn{2}{c}{Reflection}\\
               & PSNR & SSIM & PSNR & SSIM \\ \hline
Fraid et al.**~\cite{Fraid1999}& 21.99&0.714&6.48& 0.241 \\ 
Schechner et al.**~\cite{2000Schechner}& 23.42&0.655&12.40& 0.247 \\ 
Kong et al.**~\cite{kong14pami}& 18.76 & 0.402  & 12.96 & 0.271      \\ 
Yang et al.~\cite{Yang_2019_CVPR}& 25.42 & 0.780  & - & -      \\ 
\hline
Wieschollek et al.*~\cite{eccv2018/Wieschollek}& 22.15 & 0.711  & 15.93 & 0.462      \\ 
BDN*~\cite{eccv18refrmv_BDN}& 24.49 & 0.805  & 12.34 & 0.377      \\ 
Wen et al.*~\cite{Wen_2019_CVPR_Linear}& 26.62 & 0.827  & - & -      \\ 
% ERR*\cite{wei2019single_ERR} & 18.52 & 0.788  & - & -      \\ 
% Wei et al.*~\cite{wei2019single_ERR}& 29.76 & 0.893  & - & -      \\ 
Wei et al.~\cite{wei2019single_ERR}& 30.13 & 0.899  & - & -      \\ 
% Zhang et al.*~\cite{zhang2018single}& 29.52 & 0.841  & 28.35 & 0.61      \\ 
Zhang et al.~\cite{zhang2018single} & 31.91 & 0.903  & 32.02 & 0.88      \\ 

\hline
Ours      & \textbf{34.62} & \textbf{0.936}  & \textbf{33.88} & \textbf{0.907}  \\ 
Ours (3 inputs)     & 33.91 & 0.930  & 33.53 & 0.903  \\ 
\Xhline{1.0pt}
\end{tabular}
\vspace{1mm}
\caption{Quantitative results on our M-R dataset. Our method outperforms all others in PSNR and SSIM. Note methods tagged with * are evaluated with pre-trained models and tag ** stands for third-party implementation. To compare fairly with other methods~\cite{Fraid1999,Schechner1999PolarizationbasedDO,kong14pami,eccv2018/Wieschollek} which use three polarization images, we use $I_1+I_3-I_2$ to represent $I_4$ as input as a `3 inputs' version.}
\label{table:Metrics}
\end{table}

\paragraph{Experimental setup}
The experiments are mainly conducted on our M-R dataset since it is the only available raw image dataset. We select 100, 107 pairs of data as a validation set and a testing set. All data are stored in the 16-bit PNG format to avoid precision loss.
%Test scenes are separated from the others. since raw data is needed, and our M-R dataset is currently the only reflection removal dataset with raw data. Our M-R dataset has 807 sets of samples, including 600 train samples, 100 validation samples, and 107 test samples according to scenes randomly. 

Most existing works train their models in RGB space. To minimize the gap between training and testing data for these methods, we average the intensity of $I_1, I_2, I_3, I_4$ followed by gamma correction ($\gamma = 1/2.2$) before inputting to their models. Note that the domain gap between RGB images and gray images may degrade the performance of some methods. All the input images and results are saved as 16-bit PNG or NPY files to avoid accuracy loss. 

\subsection{Comparisons with baselines}
\paragraph{Quantitative evaluation} 
Table \ref{table:Metrics} summarizes the evaluation results on our dataset. Our method presents a new state-of-the-art performance. Performance of traditional polarization-based methods~\cite{2000Schechner, kong14pami, Fraid1999} rank low since their assumption that all light sources are unpolarized is oversimplified for real-world data. An interesting phenomenon is that BDN~\cite{eccv18refrmv_BDN} scores badly in reflection despite its bidirectional network design. After analysis, we find out that BDN confuses between transmission and reflection in many cases, which affects the performance significantly. Scores of Zhang et al.~\cite{zhang2018single} and Wei et al.~\cite{wei2019single_ERR} are the closest to ours. In addition to being retrained on our dataset, another common characteristic of the two methods is that they are designed for not only synthetic data but also real data. 
%The model by Yang et al.~\cite{Yang_2019_CVPR} focuses on suppressing the reflection, and its performance is in the middle of all the models. 

\begin{figure}[t]
\centering
\begin{tabular}{@{}c@{\hspace{1mm}}c@{\hspace{1mm}}c@{\hspace{1mm}}c@{}}
&\includegraphics[width=0.32\linewidth]{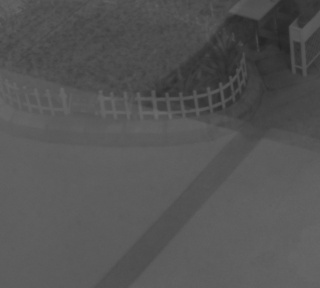}&
\includegraphics[width=0.32\linewidth]{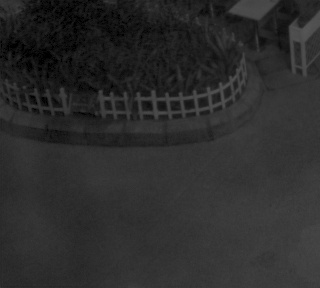}&
\includegraphics[width=0.32\linewidth]{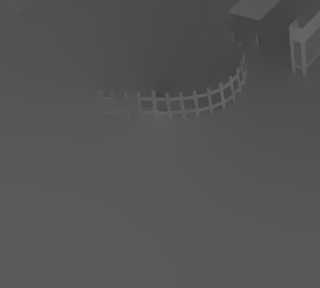}\\
&Input & Ours & Yang et al.~\cite{Yang_2019_CVPR}\\
&\includegraphics[width=0.32\linewidth]{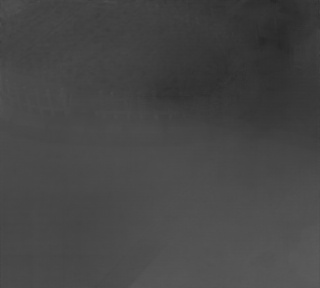}&
\includegraphics[width=0.32\linewidth]{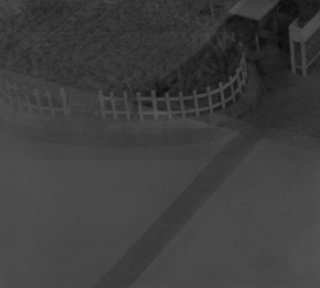}&
\includegraphics[width=0.32\linewidth]{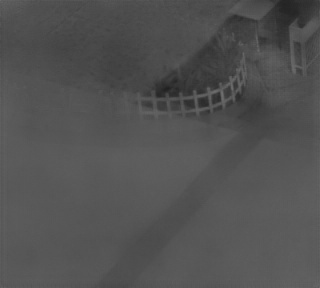}\\
&BDN~\cite{eccv18refrmv_BDN} & Wei et al.~\cite{wei2019single_ERR}& Wen et al.~\cite{Wen_2019_CVPR_Linear}\\
\end{tabular}
%\vspace{1mm}
\caption{An extreme that image is not in good focus, where the transmission is a little blurry. Previous methods tend to remove too much content since they assume that the reflection is blurry.}
\label{fig:Percep-HardCase}
\end{figure}

\begin{figure}[t]
\centering
\begin{tabular}{@{}c@{\hspace{1mm}}c@{\hspace{1mm}}c@{\hspace{1mm}}c@{}}
% \rotatebox{90}{\small \hspace{2mm}With Pol}
% \rotatebox{90}{\small \hspace{6mm}RGB}
&\includegraphics[width=0.32\linewidth]{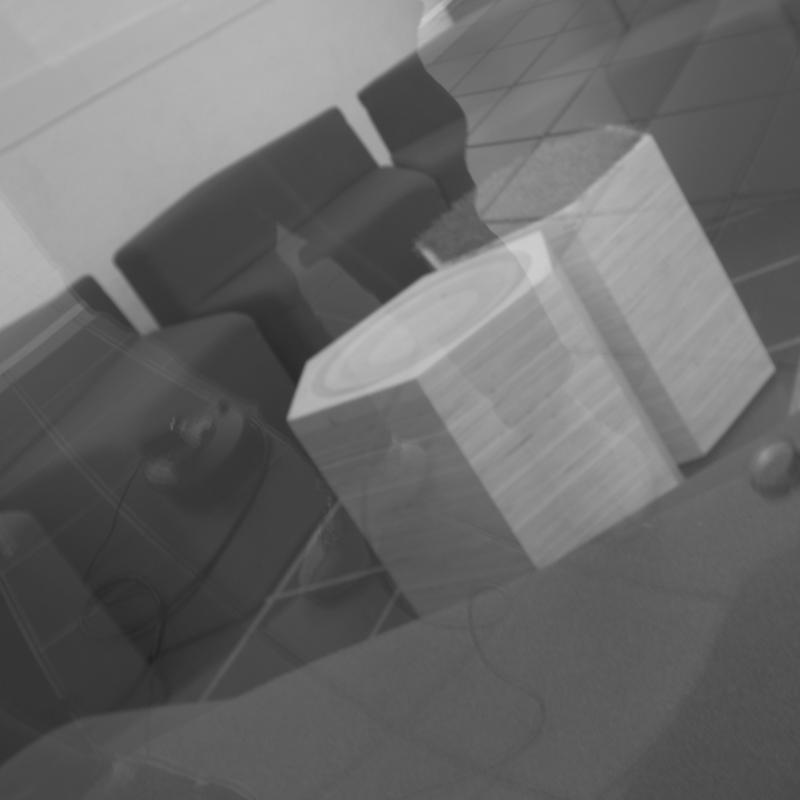}&
\includegraphics[width=0.32\linewidth]{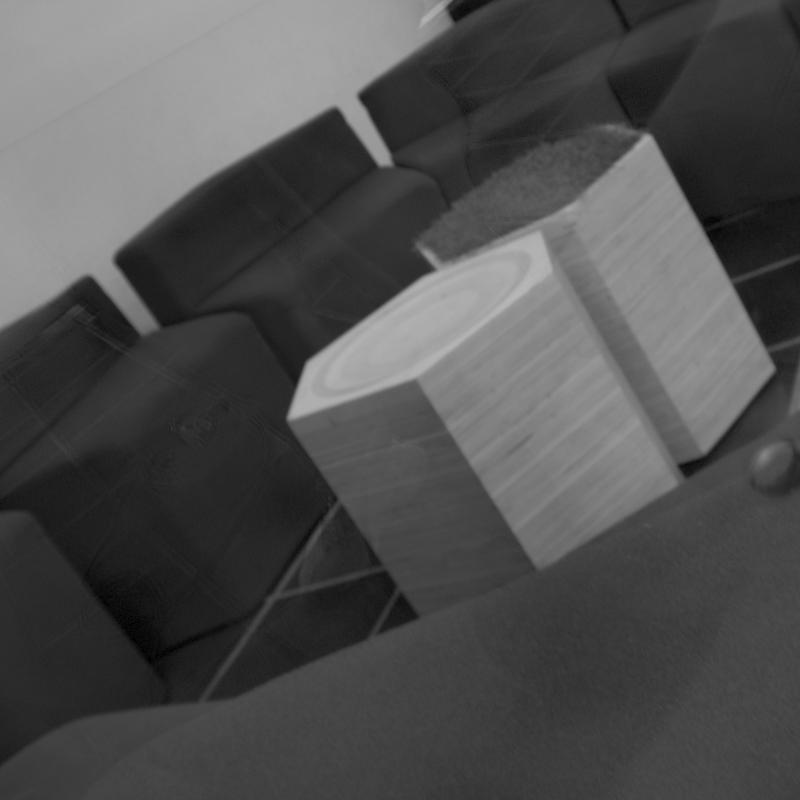}&
\includegraphics[width=0.32\linewidth]{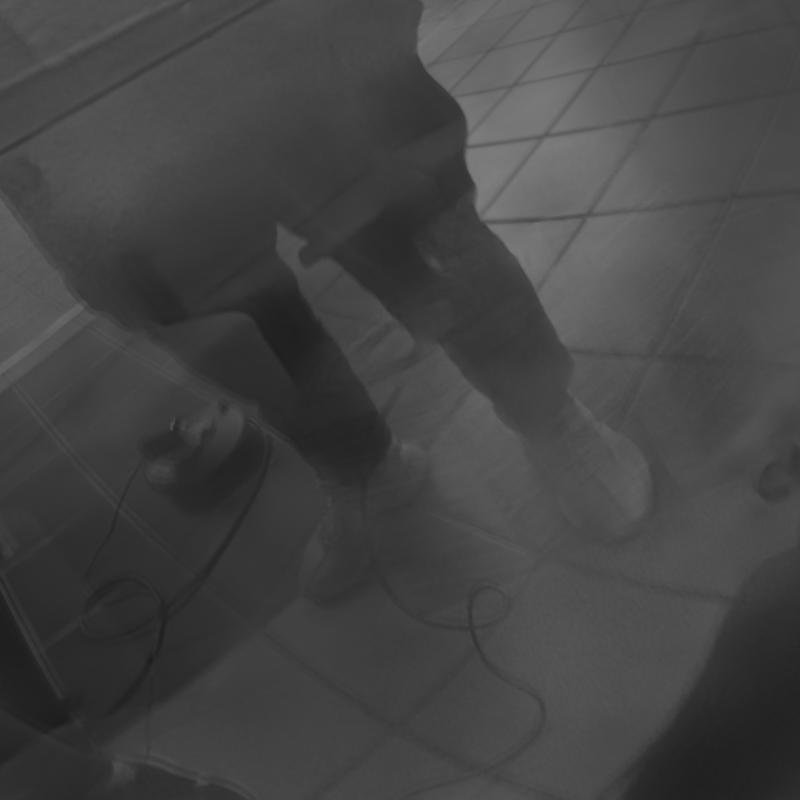}\\
% \rotatebox{90}{\small \hspace{2mm}W/o Pol}
&Input & Our $T$ & Our $R$  \\
\end{tabular}
\caption{Our method achieves satisfying results on curved glass.}
\label{fig:Curved}
\end{figure}

\begin{figure}[t]
\centering
\begin{tabular}{@{}c@{\hspace{1mm}}c@{\hspace{1mm}}c@{\hspace{1mm}}c@{}}
\includegraphics[width=0.24\linewidth]{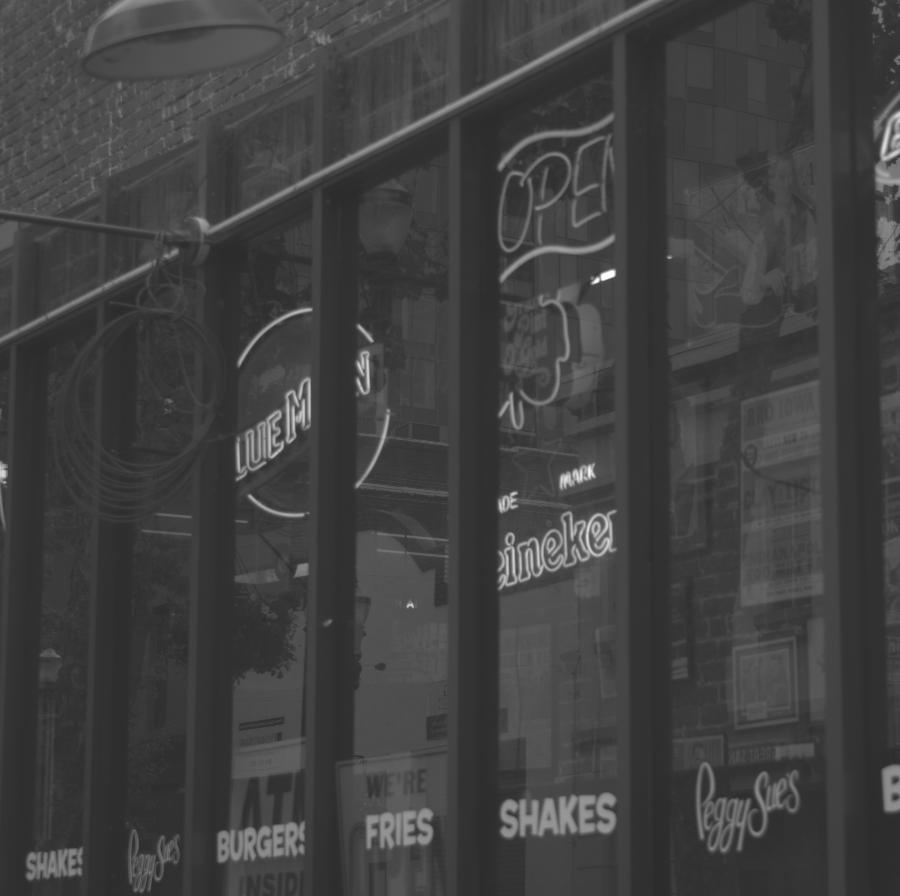}&
\includegraphics[width=0.24\linewidth]{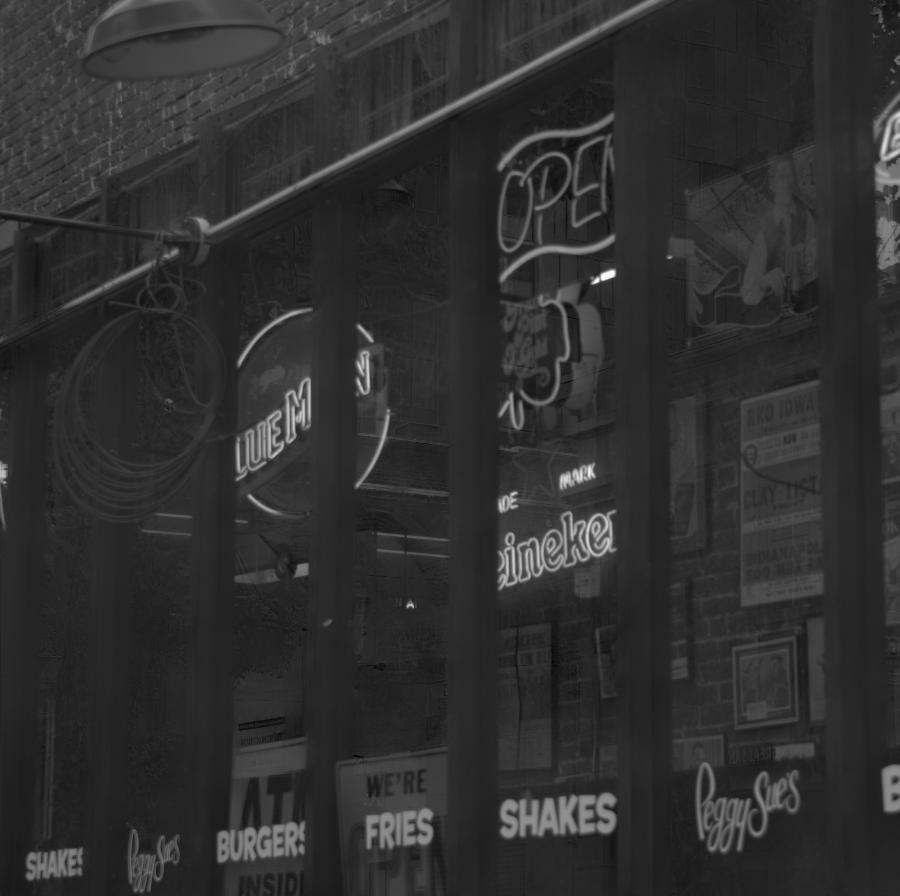}&
\includegraphics[width=0.24\linewidth]{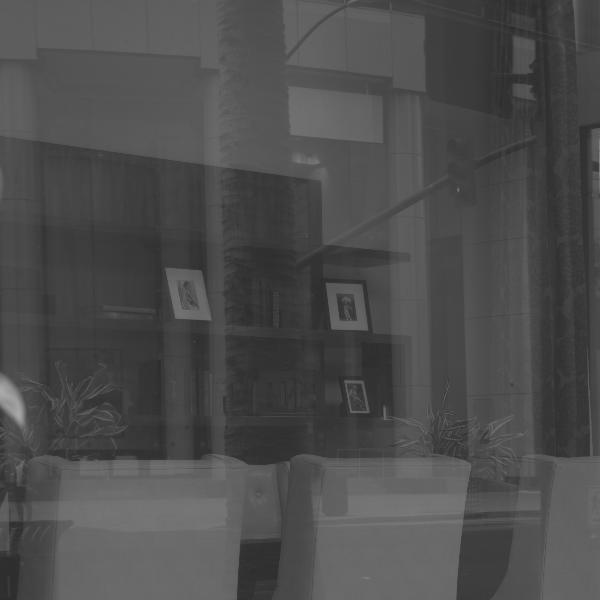}&
\includegraphics[width=0.24\linewidth]{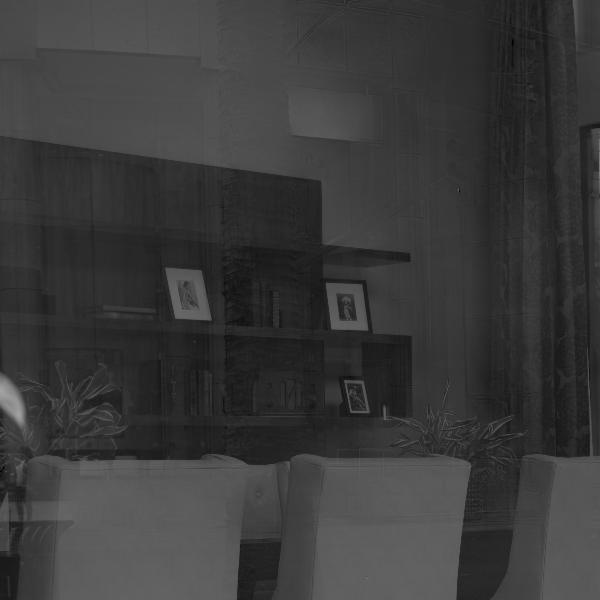}\\
% \rotatebox{90}{\small \hspace{2mm}W/o Pol}
Input & Our $T$ &Input & Our $T$  \\
\end{tabular}
\caption{Our performance is reasonable on non-ideal data~\cite{eccv2018/Wieschollek}}
\label{fig:NonIdeal}
\end{figure}

\paragraph{Qualitative evaluation}
Fig. \ref{fig:Perceptual Comparison} shows several samples by different methods in different situations. We choose the best two single image models and the best polarization method for perceptual comparisons. As seen in Fig. \ref{fig:Perceptual Comparison}, our method can handle different types of reflection well and remove the reflection pretty well without introducing artifacts. Wieschollek et al.~\cite{eccv2018/Wieschollek} can also remove different types of reflection based on polarization, but their results have visible artifacts, and it even amplifies the reflection for the third case. For Zhang et al.~\cite{zhang2018single} and Wei et al.~\cite{wei2019single_ERR}, the results have visible residual reflection left. Fig. \ref{fig:Percep-HardCase} shows a hard case where the mixed image is a bit blurry. Previous methods~\cite{Yang_2019_CVPR, eccv18refrmv_BDN, wei2019single_ERR, Wen_2019_CVPR_Linear} assuming the reflection is blurry perform poorly and tend to remove too much content. Our result shows better generalization without such an assumption. Our model can also achieve good performance on curved glass and non-ideal data collected by Wieschollek et al.~\cite{eccv2018/Wieschollek}, as shown in Fig.~\ref{fig:Curved} and Fig.~\ref{fig:NonIdeal}.
%difficult case for previous single image reflection removal methods. The methods that assume the reflection is blurry perform poorly when reflection is sharp and strong.  Also, when the mixed images are blurry, these methods tend to remove too much, including objects in transmission.
 
\subsection{Ablation study}
%We conduct an ablation study and controlled experiments to analyze the importance of each module of our model.  Table \ref{fig:Perceptual Comparison} summarizes the ablation study by conducting a perceptual user study on our M-R dataset. 

\begin{figure}[t]
\centering
\begin{tabular}{@{}c@{\hspace{1mm}}c@{\hspace{1mm}}c@{\hspace{1mm}}c@{}}
% \rotatebox{90}{\small \hspace{2mm}With Pol}
% \rotatebox{90}{\small \hspace{6mm}RGB}
&\includegraphics[width=0.32\linewidth]{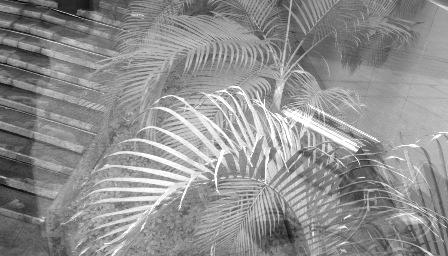}&
\includegraphics[width=0.32\linewidth]{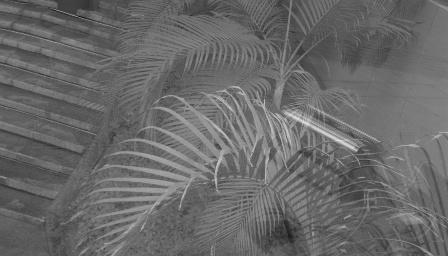}&
\includegraphics[width=0.32\linewidth]{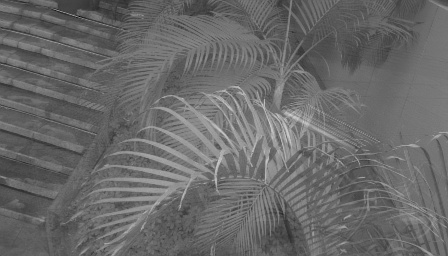}\\
&Input & $\hat T$ without pol & $\hat T$ with pol\\
% \rotatebox{90}{\small \hspace{2mm}W/o Pol}
&\includegraphics[width=0.32\linewidth]{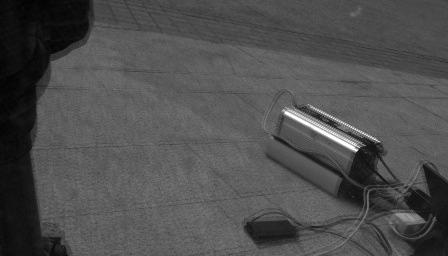}&
\includegraphics[width=0.32\linewidth]{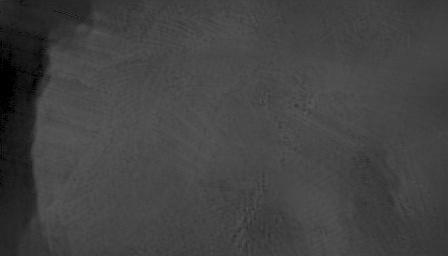}&
\includegraphics[width=0.32\linewidth]{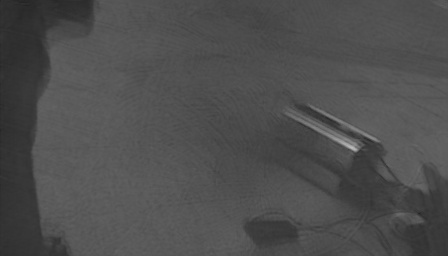}\\
&GT R & $\hat R$  without pol & $\hat R$ with pol \\
\end{tabular}
%\vspace{1mm}
\caption{Without the polarization information, the network cannot distinguish objects from $R$ or $T$ in many cases, especially for sharp reflection.}
\label{fig:With pol}
\end{figure}

\begin{table}[t]
\small
\centering
\renewcommand{\arraystretch}{1.2}
\begin{tabular}{lcccc}
\Xhline{1.0pt}
& \multicolumn{2}{c}{Transmission} & \multicolumn{2}{c}{Reflection}\\
               & PSNR & SSIM & PSNR & SSIM \\ \hline

Without polarization & 31.92 & 0.919  & 31.38 & 0.876  \\ 
Without two-stage & 32.78 & 0.920  & - & -      \\ 
Without PNCC & 34.42 & 0.934  & 33.72 & 0.905      \\ 
\hline
Ours      & \textbf{34.62} & \textbf{0.936}  & \textbf{33.88} & \textbf{0.907}  \\ 
\Xhline{1.0pt}
\end{tabular}
\vspace{1mm}
\caption{Results of the ablation study. For the model without a two-stage design, there is no reflection as we only estimate $T$. }
\vspace{-2mm}
\label{table:Ablation study}
\end{table}

%We conduct ablation study and controlled experiments to analyze the importance of each module of our model. 
To study the influence of polarization information, we replace the input channels $I_1, I_2, I_3, I_4, \rho, \phi$ all with $I$ and keep the network structure the same. To study the effect of our two-stage structure, we remove the loss on $R$. Finally, we conduct an experiment with the setting without PNCC.
The results are shown in Table \ref{table:Ablation study}. Polarization information improves the performance most. 
Fig. \ref{fig:With pol} shows a sample. The model predicts $T$ as $R$ without the support of polarization information. The two-stage design also boosts the performance of our model by a large margin. Our proposed PNCC can further increase the performance of our model on reflection removal.

As an additional evaluation, we compare PNCC with the exclusion loss proposed by Zhang et al.~\cite{zhang2018single}. The experiment is conducted in DoubleDIP~\cite{DoubleDIP} framework, which adopts exclusion loss to decompose images. By replacing the exclusion loss with our PNCC, we get the evaluation results in Table \ref{table:DIP}. Our approach outperforms their official implementation easily and still performs better after tuning the hyperparameters for them. 
% Since it takes about 10 minutes for seperating a image by DoubleDIP and no training data is required, we choose 100 images as validation set and test set.
%we apply PNCC to DoubleDIP~\cite{DoubleDIP}, an unsupervised image decomposition method. We replace the original exclusion loss with our PNCC loss. We outperform their official implementation easily. We also help their model by tuning hyperparameters to get better results. In the end, our PNCC loss function still performs better than the exclusion loss. The detailed results can be found in Table \ref{table:DIP}.

\begin{table}[t]
\small
\centering
\renewcommand{\arraystretch}{1.2}
\begin{tabular}{lccc}
\Xhline{1.0pt}
               & Exclusion & Exclusion (tuned) & PNCC \\ \hline
PSNR & 22.43 & 26.04 & \textbf{26.33}\\  \Xhline{1.0pt}
\end{tabular}
\vspace{1mm}
\caption{The performance of the exclussion loss and the PNCC loss for DoubleDIP~\cite{DoubleDIP}. We tune the hyperparameters of DoubleDIP with the exclusion loss by grid search.}
\label{table:DIP}
\end{table}

\section{Discussion}

We propose a two-stage polarized reflection removal model with perfect alignment of input-output image pairs. With a new reflection formulation to bypass the misalignment problem between the background and mixed images, we build a polarized reflection removal dataset that covers more than 100 types of glass in the real world. A general decomposition loss called PNCC is proposed to minimize the correlation of two images at different feature levels. We have conducted thorough experiments to demonstrate the effectiveness of our model. We hope our novel model formulation and the M-R dataset can inspire research in reflection removal in the future.

\section*{Acknowledgement} We thank SenseTime Group Limited for supporting this research project.

{\small
\bibliographystyle{ieee_fullname}
\bibliography{egbib}
}

\end{document}